\documentclass[10pt,twocolumn,letterpaper]{article}

\usepackage{iccv}
\usepackage{epsfig}
\usepackage{graphicx}
\usepackage{amsmath}
\usepackage{amssymb}
\usepackage{times}
\usepackage{multirow}
\usepackage{pifont}
\usepackage{booktabs}
\usepackage{booktabs,multirow}
\usepackage{graphics}
\usepackage{threeparttable}
\usepackage{color}
\usepackage[normalem]{ulem}
\usepackage{multirow}
\usepackage{float}
\usepackage{amsfonts}
\usepackage{bm}
\usepackage{enumitem}
\usepackage[numbers,sort&compress]{natbib}
\usepackage{array}
\usepackage[table]{xcolor}
\usepackage{colortbl}
\definecolor{myy}{RGB}{126,95,0}
\definecolor{mygray}{gray}{.9}
\definecolor{bblue}{RGB}{30,80,120}
\definecolor{mygray1}{gray}{.7}
\usepackage{bm}
\usepackage{mathtools}
\usepackage{subcaption}
\usepackage{siunitx}
\usepackage[nopar]{lipsum}
\usepackage[export]{adjustbox}

\usepackage{url}
\usepackage{nicefrac}       
\usepackage{microtype}      
\usepackage{algorithm}
\usepackage{algorithmic}
\usepackage{wrapfig}
\usepackage{xcolor}
\usepackage{arydshln}
\usepackage{listings}
\usepackage{dblfloatfix}

\usepackage{soul}
\soulregister\cite7
\soulregister\ref7
\soulregister\pageref7

\newcolumntype{I}{!{\vrule width 1pt}}

\definecolor{mygray}{gray}{.9}
\definecolor{ggray}{RGB}{127,127,127}
\definecolor{reda}{RGB}{192,0,0}
\definecolor{redb}{RGB}{217,148,143}
\definecolor{myyellow}{RGB}{190,144,0}
\definecolor{mygreen}{RGB}{80,100,40}
\definecolor{myblue}{RGB}{30,90,100}
\definecolor{apurple}{RGB}{102, 77, 166}
\definecolor{agreen}{RGB}{117, 175, 92}
\definecolor{ablue}{RGB}{66, 115, 147}
\definecolor{ablue}{RGB}{66, 115, 147}
\definecolor{ablack}{RGB}{57, 57, 57}
\definecolor{agray}{RGB}{101, 101, 101}
\newcommand{\tabincell}[2]{\begin{tabular}{@{}#1@{}}#2\end{tabular}}
\newcommand{\reshl}[2]{
\textbf{#1}\fontsize{7.5pt}{1em}\selectfont{\color{mygreen}{~$\uparrow$\textbf{#2}}}
}
\newcommand{\pub}[1]{{\color{gray}{\tiny{~[{#1}]~}}}}

\makeatletter
\newcommand{\thickhline}{
  \noalign {\ifnum 0=`}\fi \hrule height 1pt
  \futurelet \reserved@a \@xhline
}

\usepackage{caption}
\usepackage[pagebackref=true,breaklinks=true,letterpaper=true,colorlinks,bookmarks=false]{hyperref}

\usepackage[capitalize]{cleveref}
\crefname{section}{Sec.}{Secs.}
\Crefname{section}{Section}{Sections}
\Crefname{table}{Table}{Tables}
\crefname{table}{Tab.}{Tabs.}

\iccvfinalcopy 

\def\iccvPaperID{10067} 

\ificcvfinal\fi

\begin{document}

\title{Clustering based Point Cloud Representation Learning for 3D Analysis}

\author{
Tuo Feng\textsuperscript{1}, Wenguan Wang\textsuperscript{2}, Xiaohan Wang\textsuperscript{2}, Yi Yang\textsuperscript{2}\footnotemark[1], Qinghua Zheng\textsuperscript{3}\\
\small \textsuperscript{1} ReLER, AAII, University of Technology Sydney~~\textsuperscript{2} ReLER, CCAI, Zhejiang University~~\textsuperscript{3} Xi'an Jiaotong University\\
\small\url{https://github.com/FengZicai/Cluster3Dseg/}
}

\maketitle
\ificcvfinal\fi

\begin{abstract}
\footnotetext[1]{Corresponding author: Yi Yang.}

Point$_{\!}$ cloud$_{\!}$ analysis (such as 3D segmentation and detec- tion) is$_{\!}$ a$_{\!}$ challenging$_{\!}$ task,$_{\!}$ because$_{\!}$ of$_{\!}$ not$_{\!}$ only$_{\!}$ the$_{\!}$ irregular geometries$_{\!}$ of$_{\!}$ many$_{\!}$ millions$_{\!}$ of$_{\!}$ unordered points,$_{\!}$ but$_{\!}$ also$_{\!}$ the$_{\!}$ great$_{\!}$ variations$_{\!}$ caused$_{\!}$ by$_{\!}$  depth,$_{\!}$ viewpoint,$_{\!}$ occlusion,$_{\!}$ \etc. Current studies put much focus on the adaption~of neural networks to the complex geometries of point clouds, but are blind to a fundamental$_{\!}$ question:$_{\!}$ \textit{how$_{\!}$ to$_{\!}$ learn$_{\!}$ an$_{\!}$ appropriate$_{\!}$ point$_{\!}$ embedding$_{\!}$ space$_{\!}$ that$_{\!}$ is$_{\!}$~aware$_{\!}$ of$_{\!}$ both$_{\!}$ discriminative$_{\!}$ semantics$_{\!}$ and$_{\!}$ challenging$_{\!}$ variations?}$_{\!\!}$ As$_{\!}$ a$_{\!}$ response,$_{\!}$  we$_{\!}$ propose$_{\!}$ a clustering based supervised learning scheme for point$_{\!}$ cloud$_{\!}$ analysis. Unlike current \textit{de-facto}, \textit{scene-wise} training paradigm, our algorithm conducts \textit{within-class} clustering on the point embedding space for automatically discovering subclass patterns$_{\!}$ which$_{\!}$ are$_{\!}$ latent$_{\!}$ yet$_{\!}$ representative$_{\!}$ \textit{across$_{\!}$ scenes}.$_{\!}$ The$_{\!}$ mined$_{\!}$ patterns$_{\!}$ are,$_{\!}$ in turn, used to repaint the embedding space, so as to respect the underlying distribution of the entire training dataset and improve the robustness to the variations. Our algorithm is principled and readily pluggable to modern~point cloud segmentation networks during training, without extra overhead during testing. With various~3D network architectures (\ie, voxel-based, point-based, Transformer-based, automatically searched), our algorithm shows notable improvements on famous point cloud$_{\!}$ segmentation$_{\!}$ datasets$_{\!}$ (\ie,$_{\!}$ \textbf{2.0-2.6}\%$_{\!}$ on$_{\!}$ single-scan$_{\!}$ and$_{\!}$ \textbf{2.0-2.2}\%$_{\!}$ multi-scan$_{\!}$ of$_{\!}$ SemanticKITTI,$_{\!}$ \textbf{1.8-1.9}\%$_{\!}$ on$_{\!}$ S3DIS,$_{\!}$ in$_{\!}$ terms$_{\!}$ of$_{\!}$ mIoU).$_{\!}$ Our algorithm also demonstrates utility in 3D detection, showing \textbf{2.0-3.4}\%$_{\!}$ mAP gains on$_{\!}$ KITTI.
\end{abstract}

\section{Introduction}
During the last few years, point cloud analysis, such as 3D segmentation, has attracted increasing research effort, due to the wide applications in autonomous driving, intelligent robotics, airborne laser scanning, and virtual reality. In particular, the advances in deep learning significantly pushed forward the state-of-the-art in this field. Applying standard neural networks which are specialized for grid-like data, such as natural images, to point clouds is nontrivial, as point data are unorganized and irregular. To adapt neural networks~to~the geometries of point data, considerable effort has been made and representative achievements include:
i) \textit{projection-/voxel-based networks}$_{\!}$~\cite{tchapmi2017segcloud,wu2018squeezeseg,graham20183d,meng2019vv,milioto2019rangenet++,choy20194d,wu2019squeezesegv2,xu2020squeezesegv3,zhang2020polarnet,cortinhal2020salsanext,hu2021vmnet} that project irregular point clouds to regular representations, so$_{\!}$ that$_{\!}$ mature$_{\!}$ 2D/3D$_{\!}$ convolution$_{\!}$
can$_{\!}$ be$_{\!}$ applied$_{\!}$ for$_{\!}$ segmentation;$_{\!}$ and$_{\!}$ ii)$_{\!}$ \textit{point-based$_{\!}$ networks}$_{\!}$~\cite{engelmann2020dilated,hua2018pointwise,zhao2019pointweb,zhang2019shellnet,zhu2021cylindrical} that ingest raw point clouds directly, by using permutation-invariant operator$_{\!}$~\cite{Qi_2017_CVPR,qi2017,Wu_2019_CVPR,yang2019modeling,hu2020randla}, graph~convolution$_{\!}$~\cite{landrieu2018large}, customized~convolution$_{\!}$~\cite{wang2018deep,ummenhofer2019lagrangian,thomas2019kpconv}, or self-attention (Transformer) based architecture$_{\!}$~\cite{zhao2021point,mazur2021cloud,fan2021point}.

Nevertheless,$_{\!}$ the$_{\!}$ challenges$_{\!}$ in$_{\!}$ point$_{\!}$ cloud$_{\!}$ analysis$_{\!}$ stem$_{\!}$ not$_{\!}$ only$_{\!}$ from$_{\!}$ the$_{\!}$ intrinsic$_{\!}$~non-Euclidean$_{\!}$ nature$_{\!}$ of$_{\!}$ point data,$_{\!}$ but$_{\!}$ also$_{\!}$ from$_{\!}$ the$_{\!}$ large$_{\!}$ intra-class$_{\!}$ variations$_{\!}$ caused$_{\!}$ by depth,$_{\!}$ occlusion,$_{\!}$ viewpoint,$_{\!}$ shape,$_{\!}$ \textit{etc}.$_{\!}$ Despite$_{\!}$ various$_{\!}$ fancy point structure-aware network designs  and their~encouraging results, a fundamental issue was long  ignored: \textit{how to learn a good point~embedding space that is discriminative for semantic categorization yet robust for point data variations?}

Mitigating this issue demands a powerful learning regime that$_{\!}$ is$_{\!}$ aware$_{\!}$ of$_{\!}$ latent$_{\!}$ variation$_{\!}$ modes$_{\!}$~(or$_{\!}$ representative$_{\!}$ fine- grained$_{\!}$ patterns) -- comprehensively describing the potential structure$_{\!}$ of$_{\!}$ point$_{\!}$ data.$_{\!}$ However,$_{\!}$ in$_{\!}$ practice,$_{\!}$ it$_{\!}$ is$_{\!}$ infeasible$_{\!}$ to precisely annotate, or even roughly identify, the underlying data patterns in point clouds. This may be the reason behind the common choice that point cloud segmentation is learned as point-wise classification; any fine-grained patterns that the point data may possess are left to be `mysteriously' learned through the supervision from high-level semantic tags.

These novel insights motivate us to devise a clustering analysis based training scheme for point cloud segmentation.\\
\noindent It complements the standard supervised learning of point-wise classification with unsupervised clustering and regularization of the feature space. Specifically, clustering is con- ducted inside each labeled semantic class to automatically discover informative yet hidden subclass patterns without explicit annotation. The discovered subclass patterns essentially capture the underlying fine-grained distribution of the whole training dataset. They are then used to reshape the point embedding space, achieved by explicitly inspiring inter-subclass/-cluster discriminativeness, and reducing intra-subclass/-cluster variation. Such regularized representation space in turn facilitates the discovery of typical within-class variation modes, and benefits point recognition eventually.

Our learning algorithm enjoys several appealing advantanges:
\textbf{First}, it raises a \textit{dataset-level context-aware} training strategy.$_{\!}$ Unlike$_{\!}$ the$_{\!}$ current$_{\!}$ \textit{de-facto},$_{\!}$ scene-wise$_{\!}$ training$_{\!}$~pa- radigm, our algorithm groups point features across training\\
\noindent   scenes, and conducts clustering based representation learning. By probing the global data distribution, our algorithm encourages the highly flexible feature space to be discretized into a few distinct subcluster centers, easing the difficulty~of the final semantic classification. \textbf{Second}, it is \textit{efficient} for large-scale point cloud training. To avoid time-consuming clustering of massive point data, we opt the Sinkhorn-Knopp algorithm$_{\!}$~\cite{knight2008sinkhorn,cuturi2013sinkhorn} that solves cluster assignment using fast matrix-vector algebra$_{\!}$~\cite{asano2019self}. Moreover, to follow closely the drifting representation during network training, a momentum update strategy is adopted for online approximation of the subcluster centers.
\textbf{Third}, it is \textit{principled} enough~to~be seamlessly incorporated into the training process of any modern point cloud segmentation networks, without bringing extra computation burden or model parameters during inference.

For thorough evaluation, we approach our training algori- thm on four remarkable point cloud~segmentation models, \ie, Cylinder3D$_{\!}$~\cite{zhu2021cylindrical} (\textit{voxel}-based), KPConv$_{\!}$~\cite{thomas2019kpconv} (\textit{point}-based),  PTV1$_{\!}$~\cite{zhao2021point} (\textit{Transformer}-based), SPVNAS$_{\!}$~\cite{tang2020searching}$_{\!}$ (\textit{neural$_{\!}$ architecture$_{\!}$ search}$_{\!}$ (NAS)$_{\!}$ based),$_{\!}$ and$_{\!}$ conduct experiments on 3D point cloud segmentation for urban scenes$_{\!}$ (\ie,$_{\!}$ SemanticKITTI$_{\!}$~\cite{behley2019semantickitti}$_{\!}$ single-scan)$_{\!}$ and$_{\!}$ indoor$_{\!}$~environments$_{\!}$ (\ie,$_{\!}$ S3DIS$_{\!}$~\cite{armeni20163d})$_{\!}$ as$_{\!}$ well$_{\!}$ as$_{\!}$ 4D$_{\!}$ segmentation$_{\!}$ of point cloud sequences (\ie, SemanticKITTI~\cite{behley2019semantickitti} multi-scan). Results show that our algorithm owns \textbf{2.2-2.6}\%, \textbf{1.9-2.2}\%, \textbf{1.8}\%,$_{\!}$ and$_{\!}$ \textbf{2.0}\%$_{\!}$ mIoU$_{\!}$ gains$_{\!}$ over$_{\!}$ Cylinder3D,$_{\!}$ KPConv,$_{\!}$ PTV1,$_{\!}$ and$_{\!}$ SPVNAS,$_{\!}$ respectively.$_{\!}$ Our$_{\!}$ algorithm$_{\!}$ even$_{\!}$ promotes$_{\!}$ 3D$_{\!}$ de- tectors$_{\!}$ Second$_{\!}$~\cite{yan2018second}$_{\!}$ and$_{\!}$ PointPillar$_{\!}$~\cite{lang2019pointpillars} by  \textbf{2.7-3.4}\% and \textbf{2.0-2.2}\% mAP on KITTI~\cite{geiger2013vision}, verifying its high generality.

\vspace{-3pt}
\section{Related Work}
\vspace{-1pt}
\noindent\textbf{Deep Learning for Static Point Cloud Segmentation.}
In general, existing algorithms for single-scan point cloud segmentation can be categorized into two schools, depending~on the underlying data representation: \textbf{i)} \textit{Projection-based} methods first transform unstructured point sets to regular 2D$_{\!}$ grid \cite{tatarchenko2018tangent,wu2018squeezeseg,wu2019squeezesegv2,milioto2019rangenet++,lyu2020learning,yang2020pfcnn,cortinhal2020salsanext,xu2020squeezesegv3},$_{\!}$ or$_{\!}$ 3D$_{\!}$ voxel$_{\!}$~\cite{Maturana2015,riegler2017octnet,graham20183d,rethage2018eccv,le2018pointgrid,liu2019point,meng2019vv,zhang2020deep,zhang2020polarnet,zhu2021cylindrical}, to enable the usage of vanilla 2D/3D convolution operation. However, 2D projection based methods are likely to discard critical geometric cues and require expensive 2D-3D back-projection after 2D segmentation, yet voxel-based architectures typically suffer from significant computation and~memory$_{\!}$ usage.$_{\!}$ \textbf{ii)}$_{\!}$ \textit{Point-based}$_{\!}$ methods,$_{\!}$ pioneered$_{\!}$ by$_{\!}$ PointNet$_{\!}$~\cite{Qi_2017_CVPR,qi2017},$_{\!}$ directly$_{\!}$ learn$_{\!}$ point-wise$_{\!}$ features$_{\!}$ from$_{\!}$ raw point clouds, usually through$_{\!}$ 1)$_{\!}$ local$_{\!}$ feature$_{\!}$ pooling$_{\!}$~\cite{li2018pointcnn,li2018so,engelmann2018know,huang2018recurrent,zhang2019shellnet,zhao2019pointweb,jiang2019hierarchical,Yan_2020_CVPR,hu2020randla,fan2021scf,lian2020large},$_{\!}$ 2)$_{\!}$ graph$_{\!}$ convolution$_{\!}$~\cite{simonovsky2017dynamic,shen2018mining,wang2018local,landrieu2018large,landrieu2019point,wang2019graph,chen2019clusternet,jiang2019hierarchical,li2019deepgcns,wang2019dynamic},$_{\!}$ 3)$_{\!}$ kernel-based$_{\!}$ convolution~\cite{su2018splatnet,hua2018pointwise,lei2019octree,komarichev2019cnn,lan2019modeling,Wu_2019_CVPR,thomas2019kpconv,mao2019interpolated},$_{\!}$ and 4) attention-based aggregation~\cite{xie2018attentional,yang2019modeling,zhao2021point,mazur2021cloud,fan2021point}. Compared with projection-based approaches, point-based methods tend to be computationally efficient and are capable of preserving point-wise semantics as well as local geometries. Unfortunately, their performance in large-scale, urban scenarios is still not desirable~\cite{hu2021towards}.

\noindent\textbf{Deep Learning for Dynamic Point Cloud Segmentation.} 4D semantic~segmentation is rather difficult as point cloud sequences  are spatially irregular yet temporally ordered. Existing approaches for dynamic point cloud segmentation can be broadly classified into two groups, in terms of the spatial-temporal information fusion strategy: \textbf{i)} \textit{Early fusion based} methods~\cite{choy20194d,fan2021pstnet} directly process point cloud sequences via adapting the standard convolution to the heterogeneous characteristics of point clouds in spatial and temporal domains. In this way, they allow spatial-temporal information fusion throughout the networks. \textbf{ii)} \textit{Late fusion based} methods~\cite{duerr2020lidar,shi2020spsequencenet,zhang2020fusion,zhou2021tempnet,fan2021point} are typically built upon existing single-scan point cloud processing models for spatial information extraction, and devoted to leveraging temporal information to enrich static features and hence to boost segmentation.

Despite their dazzling network designs, existing static/dynamic point cloud segmentation models generally follow a \textit{scene-wise} training protocol, which treats each point data as an individual training sample and accumulates all the~point\\
\noindent  classification errors within each scene for network parameter optimization. As a result, they ignore the rich relations between points across different scenes, and fail to regularize the feature embedding space from a holistic view. In contrast, through automatic, \textit{class-wise} data clustering, our training algorithm grasps the latent structure of the whole training dataset, which draws on a key insight that meaningful, latent data structure, like subclass semantics, fine-grained patterns, and intra-class variation modes, are common and stable across scenes. As we will show, representation learned in such a way is desirable for detailed analysis of point clouds.

\noindent\textbf{Self-supervised$_{\!}$ Representation$_{\!}$ Learning$_{\!}$ and$_{\!}$ Clustering.$_{\!\!}$} Our algorithm relies on automated~discovery of unknown subclasses,$_{\!}$ achieved$_{\!}$ by$_{\!}$ clustering$_{\!}$ point$_{\!}$ data$_{\!}$ with$_{\!}$ only$_{\!}$ coarse-grained$_{\!}$ class$_{\!}$~labels.$_{\!}$ Thus$_{\!}$ it$_{\!}$ bears$_{\!}$ some$_{\!}$ resemblance$_{\!}$ to$_{\!}$ self-supervised learning techniques which learn meaningful representations from massive unlabeled data. A spectrum of recent unsupervised representation learning methods$_{\!}$~\cite{chen2020improved,chen2020simple,he2020momentum,yin2022proposalcontrast} build upon the \textit{instance discrimination} task that considers each data instance of the dataset$_{\!}$ as$_{\!}$ its$_{\!}$ own$_{\!}$ class$_{\!}$~\cite{dosovitskiy2015discriminative,caron2020unsupervised}.$_{\!}$ They$_{\!}$ conduct$_{\!}$ noise$_{\!}$ contrastive$_{\!}$ estimation$_{\!}$~\cite{gutmann2010noise},$_{\!}$ a$_{\!}$ special$_{\!}$~form$_{\!}$ of$_{\!}$ contrastive learning$_{\!}$~\cite{oord2018representation,hjelm2019learning}, to compare instances, and also show promise in dense 2D/3D representation learning \cite{landrieu2019point,xie2020pointcontrast,xie2021propagate,wang2021dense,wang2021exploring,zhou2022rethinking,lianggmmseg,jiang2021guided,yin2022semi,meng2021towards}. Another line of methods$_{\!}$~\cite{xie2016unsupervised,yang2016joint,caron2018deep,ji2019invariant,asano2019self,caron2020unsupervised,zhou2021group,li2020prototypical,li2021contrastive,liang2023clustseg,wang2022looking} discriminates between groups of images with similar features instead of individual images, by jointly performing unsupervised  representation learning and clustering.

\begin{figure*}[t]
  \centering
      \includegraphics[width=1 \linewidth]{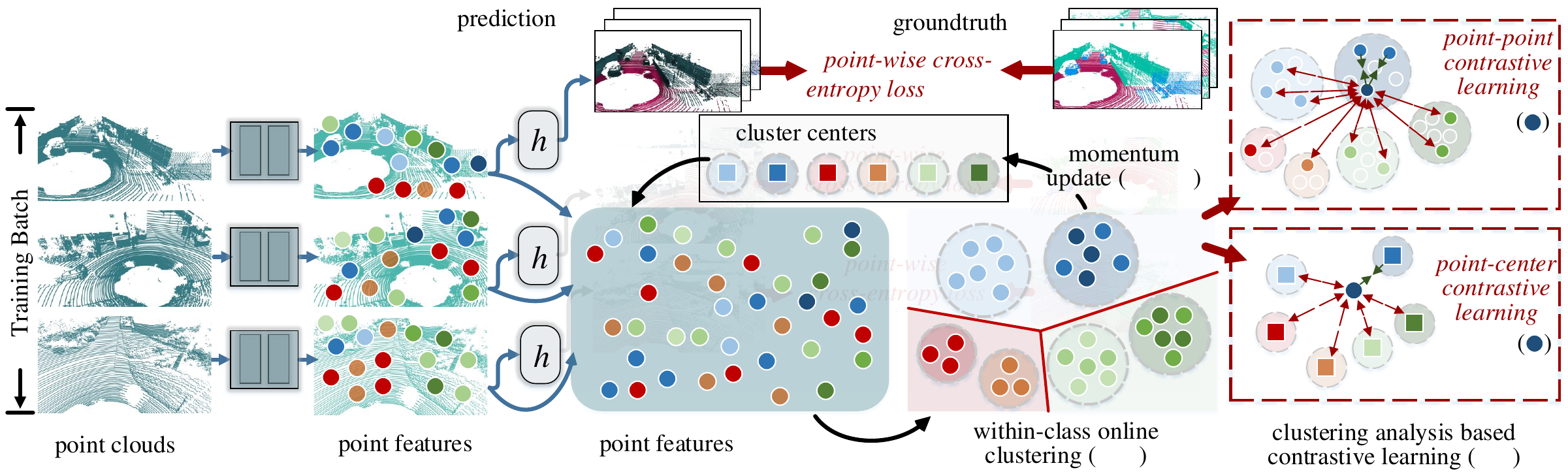}
      \put(-149,3){\scalebox{.85}{\small\S\ref{sec:3.2}}}
      \put(-29.5,3){\scalebox{.85}{\small\S\ref{sec:3.3}}}
      \put(-32,40){\scalebox{.85}{\small$\mathcal{J}_{\text{PCC}}$}}
      \put(-32,110){\scalebox{.85}{\small$\mathcal{J}_{\text{PPC}}$}}
           \put(-28,100){\scalebox{.85}{\small Eq.$_{\!}$~(\ref{eq:pNCE})}}
           \put(-28,30){\scalebox{.85}{\small Eq.$_{\!}$~(\ref{eq:pNCE2})}}
      \put(-140,91){\scalebox{.80}{Eq.$_{\!}$~(\ref{eq:update})}}
      \put(-175,141){\scalebox{.80}{$\mathcal{J}^{k\!}$} }
      \put(-202,120){\scalebox{.80}{$\mathcal{J}_{\text{CE}}$\!~(Eq.$_{\!}$~(\ref{eq:ce}))}}
      \put(-218,106){\scalebox{.80}{$\{\bm{Q}^c\}_c$ }}
      \put(-263,8){\scalebox{.80}{$\{\bm{P}^{c\!}\}_c$ }}
      \put(-319,142){\scalebox{.80}{$\mathcal{Y}^{k\!}$ }}
      \put(-416,102){\scalebox{.80}{$\varphi$ }}
      \put(-416,68){\scalebox{.80}{$\varphi$ }}
      \put(-416,37){\scalebox{.80}{$\varphi$ }}
      \put(-440,110){\scalebox{.80}{$\mathcal{P}^{k\!}$} }
      \vspace{-6pt}
\caption{$_{\!}$Overview$_{\!}$ of$_{\!}$ our$_{\!}$ clustering$_{\!}$ based$_{\!}$ supervised$_{\!}$ learning$_{\!}$ algorithm$_{\!}$ for$_{\!}$ point$_{\!}$ cloud$_{\!}$ segmentation.$_{\!\!\!}$}
\label{fig:model}
\vspace{-10pt}
\end{figure*}
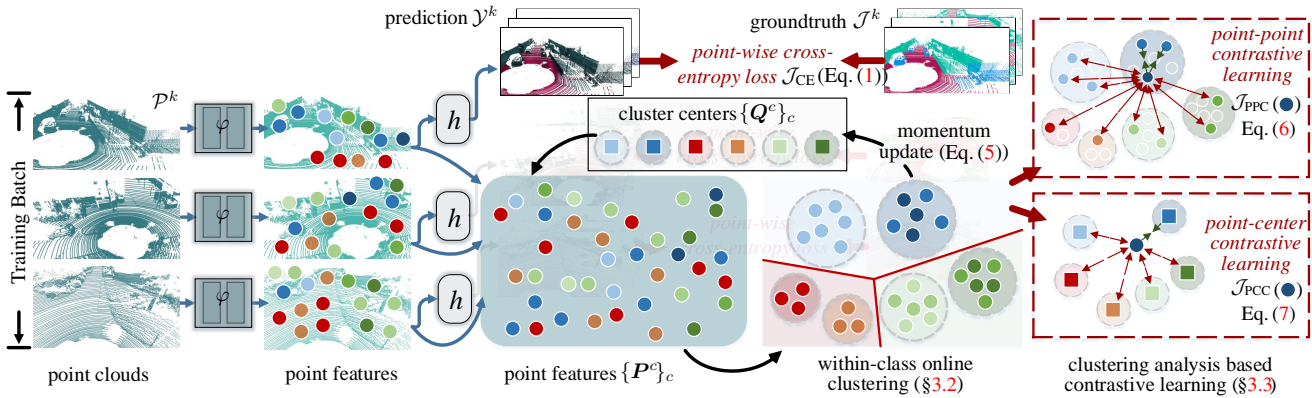

In this work, we resort to clustering to probe the under- lying structure of large-scale point sets and discover fine-grained patterns within manually-labeled, high-level seman- tic classes. We reinforce the standard supervised training paradigm of point recognition with clustering analysis based point representation learning, which regularizes the feature space by respecting the inherent structure of point data. This represents the first effort, as far as we know, that explores automatic, fine-grained pattern mining in the context of fully supervised learning of point cloud segmentation.

\vspace{-3pt}
\section{Proposed Algorithm}
\label{sec:method}
\vspace{-1pt}

\subsection{Problem Statement and Algorithm  Overview}
\label{sec:Problem}
\vspace{-1pt}
In the context of \textit{fully supervised} learning of point cloud segmentation,$_{\!}$ current$_{\!}$ common$_{\!}$ practice$_{\!}$ is$_{\!}$ to$_{\!}$ learn$_{\!}$ a$_{\!}$ point$_{\!}$ re-$_{\!}$\\
\noindent cognition$_{\!}$ network$_{\!}$ from$_{\!}$ a$_{\!}$ training$_{\!}$ dataset$_{\!}$ $\{\mathcal{P}^k,_{\!} \mathcal{L}^k\}_{k}$.$_{\!}$  Here $\mathcal{P}^{k\!}\!=\!\{p^k_{n\!}\!\in\!\mathbb{R}^{3+x}\}_{n=1\!}^{N}$ is the $k$-\textit{th} point cloud containing $N$~points with 3D position and other auxiliary information (\eg, color,$_{\!}$ intensity);$_{\!}$ $\mathcal{L}^{k\!}\!=\!\{l^k_{n\!}\!\in\!\mathcal{C}\}_{n=1\!}^{N}$ contains$_{\!}$ semantic$_{\!}$ labels$_{\!}$ for$_{\!}$ the$_{\!}$ points$_{\!}$ in$_{\!}$ $\mathcal{P}^k$,$_{\!}$ where$_{\!}$ $\mathcal{C}$ is$_{\!}$ the$_{\!}$~label$_{\!}$ list,$_{\!}$ \eg,$_{\!}$ $\mathcal{C}\!=$ $\{car, road, \cdots\}$. The segmentation network is achieved as $h_{\!}\circ_{\!}\varphi_{\!\!}:_{\!}\mathcal{P}\!\mapsto\!\mathcal{L}$, where~$\varphi_{\!}:$ $\mathbb{R}^{\!N_{\!}\times_{\!}(3+x)\!}\!\mapsto\!\mathbb{R}^{\!N_{\!}\times_{\!}d\!}$ is$_{\!}$ a$_{\!}$ \textit{feature$_{\!}$ extractor}$_{\!}$ (\includegraphics[scale=0.23,valign=c]{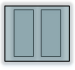}$_{\!}$ in$_{\!}$ Fig.$_{\!\!}$~\ref{fig:model})$_{\!}$ that$_{\!}$  embeds$_{\!}$ points$_{\!}$ in$_{\!}$ $\mathcal{P}$$_{\!}$ into$_{\!}$ a$_{\!}$ $d$-dimensional$_{\!}$ feature$_{\!}$ space,$_{\!}$ and$_{\!}$ $h_{\!\!}:_{\!\!} \mathbb{R}^{\!N_{\!}\times_{\!}d\!}\!\mapsto_{\!}\!\mathbb{R}^{\!N\!\times_{\!}|\mathcal{C}|\!}$~is$_{\!}$ a$_{\!}$ \textit{segmentation$_{\!}$ head}$_{\!}$ (\includegraphics[scale=0.2,valign=c]{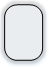}) usually consisting of a small MLP, mapping point features into the discriminative semantic space for point-wise, $|\mathcal{C}|$-way classification. Thus the whole network is typically learned by minimizing the point-wise cross-entropy loss\footnote{In$_{\!}$ practice,$_{\!}$ some$_{\!}$ other$_{\!}$ losses$_{\!}$ (\eg,$_{\!}$ lov$\acute{a}$sz loss \cite{berman2018lovasz})$_{\!}$ can$_{\!}$ be$_{\!}$ used$_{\!}$ as$_{\!}$ complementary,$_{\!}$ but$_{\!}$ this$_{\!}$ does$_{\!}$ not$_{\!}$ affect$_{\!}$ our$_{\!}$ conclusion.$_{\!\!\!}$}:\!
\vspace{-2pt}
\begin{equation}\small
\begin{aligned}\label{eq:ce}
\!\mathcal{J}_{\text{CE}}(p_n)\!=\!-\log P(l_n|p_n) \!=\!  -\log\frac{\exp(y_{n,l_n})}{\sum_{c\in\mathcal{C}\!}\exp(y_{n,c})},\!\!
\end{aligned}
\end{equation}
where $\bm{y}_{n\!}\!=\![y_{n,c}]_{c\!}\!\in\!\mathbb{R}^{|\mathcal{C}|\!}$ is the vector of categorical scores (\textit{logits}) for point $p_n$, \ie, $\bm{y}_n\!=\!h(\bm{p}_{n})$,~and $\bm{p}_{n\!}\!\in\!\mathbb{R}^d$ is the
feature of $p_n$ obtained from  $\varphi$. For the feature extractor $\varphi$, there already~have~many candidates$_{\!}$ (\eg,$_{\!}$ voxel-/point-based$_{\!}$ 3D$_{\!}$ networks)$_{\!}$ elaborately$_{\!}$ designed$_{\!}$ to$_{\!}$ capture$_{\!}$ the$_{\!}$ specific$_{\!}$ geo- metries of point data. However, point clouds yield rich and\\
\noindent diverse$_{\!}$ patterns,$_{\!}$ \eg,$_{\!}$ fine-grained$_{\!}$ semantics,$_{\!}$ intra-class$_{\!}$ varia- tions, \textit{etc}. These patterns reflect underlying data structures;

\noindent they are informative$_{\!}$ yet$_{\!}$ challenging$_{\!}$ for$_{\!}$ semantic$_{\!}$ understanding, and$_{\!}$ even$_{\!}$ hard$_{\!}$ to$_{\!}$ be$_{\!}$ identified.$_{\!}$ Thus$_{\!}$~it$_{\!}$~is$_{\!}$ usually the case that simply learning the segmentation network  $h_{\!}\circ_{\!}\varphi$ from the supervision of easily-acquired high-level semantic tags (\ie, Eq.$_{\!}$~\ref{eq:ce}), without considering the underlying data structures.

We instead devise a \textit{clustering analysis} based supervised learning framework (Fig.$_{\!}$~\ref{fig:model}). Our algorithm not only learns point recognition with pre-given semantic tags, but more essentially, it automatically discovers and encodes latent~structures of point data into the feature space $\varphi$. Features learned in such strategy are expected to be more discriminative for (fine-grained) semantics and robust for intra-class variations, hence facilitating final dense recognition of point clouds.

At each training iteration, our algorithm has two phases. In$_{\!}$ \textbf{{phase~$_{\!}$1}},$_{\!}$ we$_{\!}$ perform$_{\!}$ online$_{\!}$ clustering$_{\!}$ over$_{\!}$~massive points inside of each labeled classes. The purpose is to search for subclass patterns which are hard to be labeled yet significant across scenes. In \textbf{{phase~2}}, in addition to~optimizing the whole segmentation network $h_{\!}\circ_{\!}\varphi$ with the point-wise classification loss $\mathcal{L}_{\text{CE}}$  as usual, we leverage deterministic cluster assignments as an auxiliary constraint to shape the feature space $\varphi$. The improved features, in turn, enable more reliable within-class clustering, and eventually boost point recognition.  Independent of a certain point segmentation network, our training scheme is powerful and general.\!

\subsection{Online Clustering based Subclass Pattern~Mining$_{\!\!\!\!\!\!\!\!}$}\label{sec:3.2}
Our algorithm is built upon an intuitive insight: capturing underlying data structures can facilitate point representation learning and semantic recognition. Thus the first major question arises: \textit{how to automatically and efficiently discover~un- derlying data structures, which cannot be explicitly labeled, from massive training points?}
This motivates us to conduct unsupervised$_{\!}$ clustering$_{\!}$ inside$_{\!}$ each$_{\!}$ labeled$_{\!}$ class$_{\!}$ $c\!\in\!\mathcal{C}_{\!}$ so$_{\!}$ as$_{\!}$ to$_{\!}$ automatically$_{\!}$ mine$_{\!}$ representative$_{\!}$ yet$_{\!}$ latent$_{\!}$ subclass patterns. To scale our algorithm to millions of point data, we formulate such within-class clustering as optimal transport, which can be efficiently solved using Sinkhorn Iteration~\cite{cuturi2013sinkhorn}. In addition, to overcome the computational expensive process of cluster center computation, which requires a full epoch over the entire dataset after every update of the representation, we adopt a momentum update strategy for proceeding online clustering simultaneously with network batch training.

\begin{figure*}[t]
  \centering
      \includegraphics[width=0.99 \linewidth]{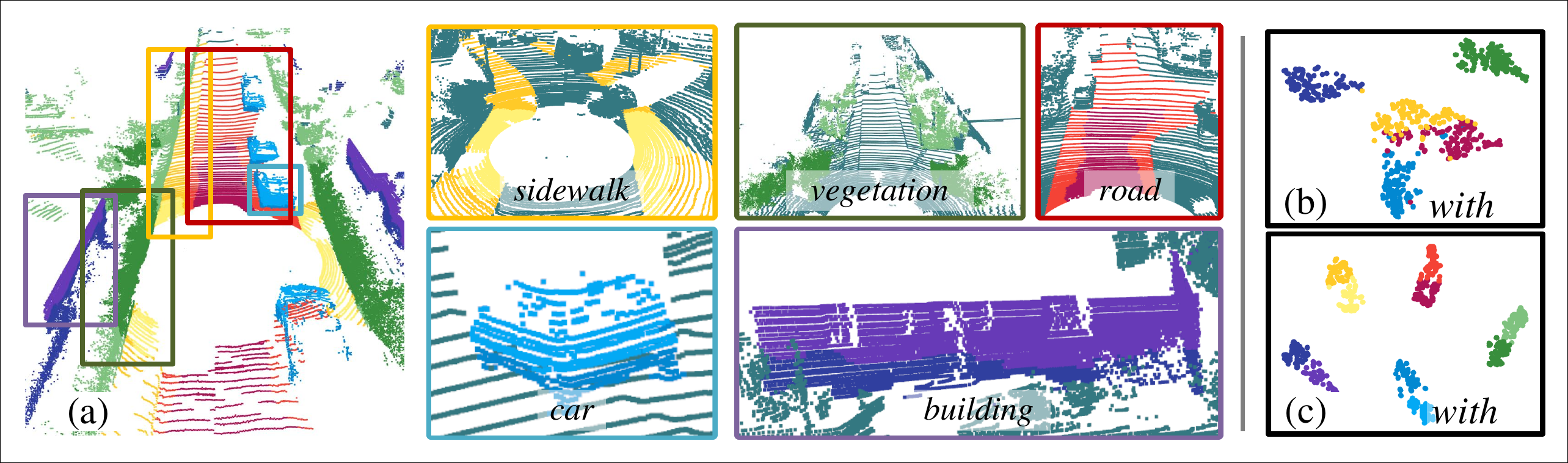}
      \put(-19.5,76){{\scalebox{.80}{$\mathcal{J}_{\text{CE}}$}}}
      \put(-17.5,10.5){{\scalebox{.80}{$\mathcal{J}$}}}
      \vspace{-6pt}
\caption{(a) Our clustering results for five classes, \ie, \textit{sidewalk}, \textit{vegetation}, \textit{road}, \textit{car}, and \textit{building}. (b-c) t-SNE visualization of point features $\{_{\!}\bm{P}^{c\!}\}_c$ learned with $\mathcal{J}_{\text{CE}}$ (Eq.$_{\!}$~(\ref{eq:update})) and $\mathcal{J}$ (Eq.$_{\!}$~(\ref{eq:overalllos})). We set $M\!=\!2$ here, see supplementary for analysis.}
\vspace{-15pt}
\label{fig:featurespace}
\end{figure*}

For each class $c_{\!}\in_{\!}\mathcal{C}$, we assume it contains $M$ latent,~fine-grained patterns. Hence there are a total of $M\!\times\!|\mathcal{C}|$ unobservable patterns are desired to be discovered from the training dataset$_{\!}$ $\{\mathcal{P}^k,\! \mathcal{L}^k\}_{k}$.$_{\!}$ To$_{\!}$ do$_{\!}$ so,$_{\!}$ we$_{\!}$ perform$_{\!}$ within-class$_{\!}$ clustering on the point embedding space $\varphi$. As a result, the training\\
\noindent  points belonging to class $c$, \ie, $\mathcal{P}^c\!=\!\{p_n|l_n\!=\!c\}$,~are~parti- tioned$_{\!}$ into$_{\!}$ $M_{\!}$ subclasses,$_{\!}$ and$_{\!}$ the$_{\!}$ $M_{\!}$ patterns$_{\!}$ of$_{\!}$ class$_{\!}$ $c_{\!}$ can$_{\!}$ be intuitively represented as the corresponding cluster centers. Let$_{\!}$ $\bm{Q}^{c\!}\!=_{\!}\![\bm{q}^c_1,_{\!} \cdots_{\!},_{\!} \bm{q}^c_M]\!\in\!\mathbb{R}^{d_{\!}\times_{\!}M\!\!}$ denote$_{\!}$ the$_{\!}$ $M_{\!}$ cluster$_{\!}$ centers$_{\!}$ of$_{\!}$ class$_{\!}$ $c$$_{\!}$ (\eg,$_{\!}$ \includegraphics[scale=0.3,valign=c]{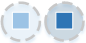}$_{\!}$ in$_{\!}$ Fig.$_{\!}$~\ref{fig:model}), and $\bm{P}^c\!=\!$ $[\bm{p}^c_1, \cdots_{\!}, \bm{p}^c_{N^c}]\!\in\!\mathbb{R}^{d_{\!}\times_{\!}N^c\!\!}$ all the$_{\!}$ features\footnote{Point feature has been projected to the unit sphere: $\bm{p}\!=\!\bm{p}/||\bm{p}||_2$; $\bm{p}$ is reused without causing ambiguity.}$_{\!}$ of$_{\!}$ points$_{\!}$ belonging$_{\!}$ to$_{\!}$ class$_{\!}$ $c$ (\eg, \includegraphics[scale=0.3,valign=c]{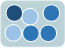}),$_{\!}$ where$_{\!}$ $p^{c\!}\!\in\!\mathcal{P}^{c\!}$ and$_{\!}$ $N^c\!=\!|\mathcal{P}^{c}|$.$_{\!}$ The$_{\!}$ cluster$_{\!}$ assignment$_{\!}$ can$_{\!}$ be represented as a binary matrix, $\bm{A}^c\!\in\!\{0,1\}^{M\!\times\!N^c}$, where~the $(m,i)$-\textit{th} element of $\bm{A}^{c\!}$ indicates whether assigning the $i$-\textit{th} point of $\mathcal{P}^{c\!}$ to the $m$-\textit{th} cluster center, \ie, the $m$-\textit{th} subclass, of $c$. The clustering inside class $c$ can be achieved as the optimization of the assignment matrix $\bm{A}^{c\!}$, \ie, maximizing the similarity between the point features and cluster centers:
\begin{equation}\small
  \begin{gathered}\label{eq:nc1}
    \!\!\min_{\bm{A}^{c}\in\mathcal{A}^{c}}\!\langle\bm{A}^{c\top\!},-\log\bm{S}^c\rangle, \\
    \!\!\!\!\mathcal{A}^{c}\!=\!\{\bm{A}^{c\!}\!\in\!\{0,1\}^{M\!\times\!N^c}|\bm{A}^{c\top}\!\bm{1}_M\!=\!\bm{1}_{N^c}, \bm{A}^{c}\bm{1}_{N^c}\!=\!\frac{N^c}{M}\bm{1}_M\}\!\!\!
  \end{gathered}
\end{equation}
where $\bm{S}^c\!=\!\text{softmax}(\bm{Q}^{c\top\!}\bm{P}^c)$ refers to the similarity matrix between cluster centers and points, $\langle \cdot \rangle$ is the Frobenius dot-product, $\log$ is applied element-wise, and
$\bm{1}_M$ denotes the vector of ones in dimension $M$. For the solution space $\mathcal{A}^{c}$, the former constraint enforces that each point is assigned~to exactly one subclass, and the later imposes an equipartition constraint~\cite{asano2019self,caron2020unsupervised} to inspire the~$N^c$ points to be grouped into $M$ subclasses of equal size. The equipartition constraint helps avoid the degenerate solution where all the point samples are partitioned to a single cluster~\cite{caron2018deep,wang2021exploring}. By relaxing $\bm{A}^{c}$ to be~an element$_{\!}$ of$_{\!}$ \textit{transportation$_{\!}$ polytope}$_{\!}$~\cite{cuturi2013sinkhorn},$_{\!}$ \ie,$_{\!}$ $\mathcal{A}'^{c}\!=\!\{\bm{A}^{c\!}\!\in\!\mathbb{R}^{M\!\times\!N^c\!}_+|\bm{A}^{c\top}\!\bm{1}_{M\!}\!=_{\!}\!\frac{1}{N^c}\bm{1}_{N^c}, \bm{A}^{c}\bm{1}_{N^c\!}\!=_{\!}\!\frac{1}{M}\bm{1}_{M\!}\}$, the label assignment task can be viewed as an instance of the \textit{optimal transport} problem, which can be efficiently solved by a fast version of the Sinkhorn-Knopp
algorithm~\cite{cuturi2013sinkhorn}:
\begin{equation}\small
\begin{aligned}\label{eq:nc12}
\!\!\!\!\min_{\bm{A}^{c}\in\mathcal{A}'^{c}}\!\langle\bm{A}^{c\top\!},-\log\bm{S}^c\!\rangle+\frac{1}{\lambda} \mathrm{KL}(\bm{A}^{c}||\frac{1}{MN^c}\bm{1}_M\bm{1}^{\top}_{N^c}),
\end{aligned}
\end{equation}
where $\mathrm{KL}$ is the Kullback-Leibler divergence, and $\lambda$ is the strength of the regularization. The solution of Prob.~(\ref{eq:nc12}) over the set $\mathcal{A}'^{c}$ can be written as:
\begin{equation}\small
  \begin{aligned}\label{eq:nc3}
    {\bm{A}}^{c*}  = \text{diag}(\bm{u})(\bm{S}^c)^\lambda\text{diag}(\bm{v}),
  \end{aligned}
\end{equation}
where exponentiation is meant element-wise. $\bm{u}\!\in\!\mathbb{R}^{M\!}$ and $\bm{v}\!\in\!\mathbb{R}^{N^c\!\!}$ are two vectors of scaling~coefficients, obtained using a small number of matrix-vector multiplications via iterative Sinkhorn-Knopp algorithm~\cite{cuturi2013sinkhorn}.$_{\!}$ Due to the drift of the point representation caused by iterative network training, after each$_{\!}$ training$_{\!}$ batch$_{\!}$ of$_{\!}$ point$_{\!}$ clouds,$_{\!}$ re-computing$_{\!}$ the$_{\!}$ cluster$_{\!}$ assignment$_{\!}$ would$_{\!}$ cost$_{\!}$ a$_{\!}$ pass$_{\!}$ over$_{\!}$ the$_{\!}$~full data. To circumvent such computationally expensive procedure of offline cluster assignment, we~restrict the transportation polytope to the minibatch, through approximating the cluster centers $\bm{Q}^c$ with momentum. As in~\cite{wang2021exploring}, at each training iteration, each cluster center $\bm{q}^{c}_m$ of class $c$ is updated as:
\begin{equation}\small
\begin{aligned}\label{eq:update}
\bm{q}^{c}_m \leftarrow  \mu\bm{q}^{c}_m + (1-\mu)\bar{\bm{p}}^{c}_m,
\end{aligned}
\end{equation}
where $\mu\!\in\![0,1]$ is the momentum coefficient, and $\bar{\bm{p}}^{c}_{m\!}$ indicates mean feature vector of the points that are assigned to cluster center $\bm{q}^{c}_m$ in the current batch. The cluster centers are initialized randomly~and gradually$_{\!}$ updated$_{\!}$ every$_{\!}$ batch,$_{\!}$ following$_{\!}$ smoothly$_{\!}$ the$_{\!}$ changing$_{\!}$ of$_{\!}$ the$_{\!}$ representation$_{\!}$ $\varphi$.$_{\!}$ These$_{\!}$ designs lead to scalable and online clustering, allowing to automatically mine latent subclass patterns from massive point data.  The clustering is very efficient on GPU; in practice, assigning 50K points into~40 clusters takes only 60 ms. We visualize clustering results ($M\!=\!2$) of five classes in Fig.~\!\ref{fig:featurespace}, where~subclasses under the same class are associated with similar colors. As seen, points with similar patterns are grouped together, thus the underlying data distribution of each class can be comprehensively captured.

\subsection{Clustering Analysis based Point Cloud Representation Learning}\label{sec:3.3}
$_{\!}$Through$_{\!}$ within-class$_{\!}$ clustering,$_{\!}$ we$_{\!}$ search$_{\!}$ for$_{\!}$ latent$_{\!}$ structures in point clouds, and detect locally~discriminative pat- terns,$_{\!}$ \ie,$_{\!}$ the$_{\!}$ cluster$_{\!}$ centers$_{\!}$ $\{\bm{q}^c_m\}_{m,c}$.$_{\!}$ The$_{\!}$ next$_{\!}$ question$_{\!}$~is:$_{\!}$ \textit{how$_{\!}$ to$_{\!}$ leverage$_{\!}$ these$_{\!}$~fine-grained$_{\!}$ patterns$_{\!}$ to$_{\!}$ aid$_{\!}$ point$_{\!}$ cloud$_{\!}$ representation learning?} To answer this, we complement the supervised point-wise classification loss $\mathcal{J}_{\text{CE}}$$_{\!}$ (Eq.$_{\!}$~(\ref{eq:ce})) with a clustering analysis based
contrastive learning strategy, which poses structured and direct supervision for point representation. In particular, with the deterministic cluster assignments in \S\ref{sec:3.2}, we conduct~contrastive representation learning over both point-point and point-center pairs. This allows us to fully exploit~relations between any two points and local data structures, and directly optimize the point feature space~$\varphi$.

\noindent\textbf{Point-Point$_{\!}$ Contrastive$_{\!}$ Learning.$_{\!}$} Our$_{\!}$ point-point$_{\!}$ contras- tive$_{\!}$ learning$_{\!}$ is$_{\!}$ achieved$_{\!}$ by$_{\!}$ comparing$_{\!}$ pairs of points to push away point representations from different subclasses while pulling~together those from the same subclass. The corres- ponding training objective for each point $p_n$ is defined as:
\begin{equation}\footnotesize\label{eq:pNCE}
\!\!\!\!\!\!\mathcal{J}_{\text{PPC}}(p_n)\!=\!\frac{1}{|\mathcal{O}_{p_n\!}|}\!\!\sum_{p^+\in\mathcal{O}_{p_n}\!\!}\!\!\!\!\!-_{\!}\log\frac{\exp(\bm{p}_{n\!}\!\cdot\!\bm{p}^{_{\!}+\!\!}/\tau)}{\exp(\bm{p}_{n\!}\!\cdot\!\bm{p}^{+\!\!}/\tau)
\!+\!\!\!\!\!\!\!\!\sum\limits_{p^{-\!}\in\mathcal{N}_{p_n}\!}\!\!\!\!\!\!\!\exp(\bm{p}_{n\!}\!\cdot\!\bm{p}^{-\!}/\tau)},
\end{equation}
where$_{\!}$ $\tau\!>\!0_{\!}$ is$_{\!}$ a$_{\!}$ scalar$_{\!}$ temperature$_{\!}$ parameter,$_{\!}$ $\mathcal{O}_{p_n\!\!}$ and $\mathcal{N}_{p_n\!\!}$ denote$_{\!}$ collections$_{\!}$~of$_{\!}$ \textit{positive}$_{\!}$ and$_{\!}$ \textit{negative} samples, respectively, for $p_n$. Training points belonging to the same cluster of $p_n$ are positive samples, while being assigned to other clusters are negative. Note that the positive (negative) samples are~not limited to a same training point cloud. To further boost our point-point contrastive learning, we follow the common practice in unsupervised representation learning~\cite{wang2020cross,chen2020improved,he2020momentum} to build a memory bank per cluster, leading to $M\!\times\!|\mathcal{C}|$ memory banks totally. The memory banks gather point features~of~corresponding clusters from previous training batches, hence increasing the quantity and diversity of~positive and negative samples. These designs deliver a \textit{cross-scene} training scheme, rather than the current \textit{de facto} scene-wise training paradigm that ignores the rich correspondences among points across different scenes.
Minimizing~Eq.~(\ref{eq:pNCE}) leads to a well-structured embedding space $\varphi$, where points with similar patterns are grouped close to each other while points with dissimilar patterns are separated.

\noindent\textbf{Point-Center Contrastive Learning.} With a similar spirit of point-point contrastive learning, \ie, inspiring intra-cluster compactness and inter-cluster separation, our point-center contrastive learning strategy contrasts the similarities between points and cluster centers on the embedding space~$\varphi$:
\begin{equation}\small\label{eq:pNCE2}
\!\!\!\!\mathcal{J}_{\text{PCC}}(p_n)\!=\!-_{\!}\log\frac{\exp(\bm{p}_n\!\cdot\!\bm{q}^{+\!\!}/\tau)}{\sum\nolimits_{c,m}\!\exp(\bm{p}_n\!\cdot\!\bm{q}^c_m/\tau)},\!\!
\end{equation}
where $\bm{q}^{+\!}$ refers to the cluster center of point $p_n$. Eq.~(\ref{eq:pNCE2}) lets $p_n$ find out the assigned cluster center~$\bm{q}^{+}$ from all the centers $\{\bm{q}^c_m\}_{c,m}$, so as to decrease the distance between $\bm{p}_n$ and~$\bm{q}^{+}$, while increasing~the distance between $\bm{p}_n$ and other cluster centers. Since cluster centers are representative of the dataset, Eq.~(\ref{eq:pNCE2}) provides a cheaper and more direct way to impose dataset-level context, or underlying data structures, on feature space optimization, compared with the point-point contrastive learning (Eq.~(\ref{eq:pNCE})). In practice, we find combining the two cluster-analysis based contrastive learning strategies yields the best performance (see detailed experiments in \S\ref{sec:ablation}). One may also view point-center contrastive learning from an
\textit{information bottleneck} perspective~\cite{tishby2015deep,ji2019invariant}, wherein the deterministic clustering imposes a natural bottleneck and discretizes the embedding space $\varphi$ as a finite set of cluster centers, \ie, $\{\bm{q}^c_m\}_{c,m}$, through minimizing Eq.$_{\!}$~(\ref{eq:pNCE2}), as opposed to learning $\varphi$ as a continuous vector space.

\noindent\textbf{Overall Training Objective.} The standard cross-entropy loss $\mathcal{J}_{\text{CE}}$ in Eq.$_{\!}$~(\ref{eq:ce}) is essentially a unary training objective that is only aware of point-wise semantic discrimination, without accounting~for any underlying data structure and pairwise relations between training points. The clustering analysis based contrastive losses, \ie,  $\mathcal{J}_{\text{PPC}}$ in Eq.~(\ref{eq:pNCE}) and $\mathcal{J}_{\text{PCC}}$ in Eq.~(\ref{eq:pNCE2}), are pairwise training objectives that exploit locally representative patterns for structure-aware, distance based point representation learning. Thus we assemble these two complementary training targets as our overall learning objective:
\begin{equation}\small\label{eq:overalllos}
\begin{aligned}
\mathcal{J}=\mathcal{J}_{\text{CE}}+\alpha(\mathcal{J}_{\text{PPC}}+\mathcal{J}_{\text{PCC}}).\!\!
\end{aligned}
\end{equation}
Our training algorithm alternately performs within-class clustering over the point embedding space~$\varphi$, and optimizes the whole segmentation network $h_{\!}\circ_{\!}\varphi$ with the semantic labels $\{\mathcal{L}^k\}_k$ and cluster assignments $\{\bm{A}^c\}_c$. As such, meaningful clusters capture fine-grained data structures and become informative supervisory signals for point representation learning; in turn, discriminative representations help obtain meaningful clusters and eventually ease point recognition.  In Fig.$_{\!}$~\ref{fig:featurespace} (b-c), we provide visualization of point embeddings learned by $\mathcal{J}_{\text{CE}}$ and $\mathcal{J}$. As seen, after additionally considering clustering analysis based training targets, the point embedding space becomes more structured.

\vspace{-1pt}
\subsection{Algorithm Details}
\label{sec:implementation}
\vspace{-1pt}
\noindent\textbf{Online$_{\!}$ Clustering$_{\!}$ (\S\ref{sec:3.2}).} We group point samples of each class$_{\!}$ into$_{\!}$ $M$$_{\!}$ subclasses$_{\!}$ for$_{\!}$ exploiting$_{\!}$ latent$_{\!}$ structures$_{\!}$ of$_{\!}$ the$_{\!}$ entire$_{\!}$ dataset. We empirically set $M\!=\!40$ and the momentum coefficient in Eq.~\!(\ref{eq:update}) $\mu\!=\!0.9999$ (related$_{\!}$ experiments$_{\!}$ can$_{\!}$ be$_{\!}$ found$_{\!}$ in$_{\!}$ \S\ref{sec:ablation}). Following~\cite{asano2019self}, we set $\lambda\!=\!25$ in Eq.~\!(\ref{eq:nc12}).

\noindent\textbf{Clustering Analysis based Training (\S\ref{sec:3.3}).} Our clustering analysis based training strategy enforces the point feature space to better respect the discovered data structures. Following the common practice in contrastive learning~\cite{li2020prototypical,chen2020simple}, we set the scalar temperature $\tau$ in Eqs.~(\ref{eq:pNCE}-\ref{eq:pNCE2}) as 0.1. For the cluster-wise memory bank, we sample $10$ point features per-cluster from each scene and store all the sampled features of all the training point clouds $\{\mathcal{P}^k\}_k$. For the training loss $\mathcal{J}$ (Eq.~(\ref{eq:overalllos})), the coefficient is set as $\alpha\!=\!1$ (we empirically find our algorithm is insensitive to $\alpha$ when $\alpha\!\in\![0,1]$).

\noindent\textbf{Point Cloud Segmentation Network $h_{\!}\circ_{\!}\varphi$.} Our algorithm is a general supervised learning scheme for point cloud segmentation. In principle, it can be applied to any segmentation networks that can learn point-wise features. In our experiments, we approach our algorithm on four typical segmentation networks, including voxel-based$_{\!}$~\cite{zhu2021cylindrical}, point-based$_{\!}$~\cite{thomas2019kpconv}, Transformer-based$_{\!}$~\cite{zhao2021point}, and NAS-based$_{\!}$~\cite{tang2020searching}.

\noindent\textbf{Inference.} Our training algorithm does not cause extra inference cost or network architectural modification during model deployment. The $M\!\times\!|\mathcal{C}|$ cluster centers and $M\!\times\!|\mathcal{C}|$ memory banks are directly discarded after network training. 

\vspace{-3pt}
\section{Experiment}
\label{sec:exp}
\vspace{-1pt}
We first report our 3D segmentation results on static point clouds of urban scenes and indoor~environments in~\S\ref{sec:ex1}~and \S\ref{sec:ex2}, respectively. Then we assess our performance on 4D segmentation of outdoor point cloud sequences in~\S\ref{sec:ex3}. For thorough evaluation, in \S\ref{sec:detection}, we extend our algorithm to 3D object detection setting and conduct experiments.  The hyperparameters mentioned in \S\ref{sec:implementation} are used for all the above experiments. Finally, in \S\ref{sec:ablation}, we provide ablative analyses on the core components of our training algorithm.

\noindent\textbf{Base$_{\!}$ Segmentation$_{\!}$ Networks.$_{\!}$} For$_{\!}$ thorough$_{\!}$ examination,$_{\!}$ we$_{\!}$ apply$_{\!}$ our$_{\!}$ training$_{\!}$ algorithm$_{\!}$ to$_{\!}$ Cylinder3D$_{\!}$~\cite{zhu2021cylindrical} (\textit{voxel}-based), KPConv$_{\!}$~\cite{thomas2019kpconv} (\textit{point}-based),
PTV1$_{\!}$~\cite{zhao2021point} (\textit{Transformer}-based), and  SPVNAS$_{\!}$~\cite{tang2020searching} (\textit{NAS}-based), which are representative for current mainstream network architectures in point cloud segmentation  and with publicly accessible implementations.$_{\!}$ For fair comparison, we adopt their default implementation settings, including hyper-parameters and augmentation recipes.

\vspace{-3pt}
\subsection{3D Segmentation on Static Urban Point Clouds}
\label{sec:ex1}
\vspace{-1pt}
\noindent\textbf{Dataset.} SemanticKITTI~\cite{behley2019semantickitti} is a large-scale driving-scene dataset$_{\!}$  for$_{\!}$  point$_{\!}$  cloud$_{\!}$  segmentation.$_{\!}$  It$_{\!}$  has$_{\!}$  $43,\!000$$_{\!}$  scans$_{\!}$  with point-wise annotation, collected from 22 sequences. Accor- ding to the official setting, we use sequences 00 to 10 for \texttt{train} (but 08 is left for \texttt{val}), and 11 to 21 for \texttt{test}. In single-scan challenge for static segmentation, 19 classes are used and mean intersection-over-union (mIoU) is reported.

\begin{table}[t]
\begin{center}
    \captionsetup{width=.48\textwidth}
\caption{$_{\!}$\textbf{Quantitative$_{\!}$ 3D$_{\!}$ segmentation$_{\!}$ results$_{\!}$} on SemanticKITTI \cite{behley2019semantickitti} single-scan challenge \texttt{test} (\S\ref{sec:ex1}). For clarity, IoUs on 6 of 19 classes are given {\small(c$_1$: \textit{sidewalk}, c$_2$: \textit{parking}, c$_3$: \textit{building}, c$_4$: \textit{truck}, c$_5$: \textit{bicycle}, c$_6$: \textit{motorcyclist})}.
}
\vspace{-2pt}
\label{table:StaticUrbantest}
\setlength\tabcolsep{3pt}
\renewcommand\arraystretch{1.05}
\resizebox{\linewidth}{!}{
\begin{tabular}{|r||c|cccccc|}
\thickhline
\rowcolor{mygray}
Method&{{mIoU}}(\%)&c$_1$(\%)&c$_2$(\%)&c$_3$(\%)&c$_4$(\%)&c$_5$(\%)&c$_6$(\%)\\
\hline
\hline
PointASNL\pub{CVPR20}\cite{Yan_2020_CVPR}&46.8&74.3&24.3&83.1&39.0&0.0&0.0\\
PolarNet\pub{CVPR20}\cite{zhang2020polarnet}&54.3&74.4&61.7&90.0&22.9&40.3&5.6\\
RandLA-Net\pub{CVPR20}\cite{hu2020randla}&55.9&74.0&61.8&89.7&43.9&29.8&9.4\\
SqueezeSegV3\pub{ECCV20}\cite{xu2020squeezesegv3}&55.9&74.8&63.4&89.0&29.6&38.7&20.1\\
SalsaNext\pub{ISVC20}\cite{cortinhal2020salsanext}&59.5&75.8&63.7&90.2&38.9&48.3&19.4\\
FusionNet\pub{ECCV20}\cite{zhang2020deep}&61.3&77.1&68.8&92.5&41.8&47.5&11.9\\
JS3C-Net\pub{AAAI21}\cite{yan2021sparse}&66.0&72.1&61.9&92.5&54.3&59.3&39.9\\
AF2S3Net\pub{CVPR21}\cite{cheng20212}&69.7&72.5&68.8&87.9&39.2&65.4&74.3\\
RPVNet\pub{ICCV21}\cite{xu2021rpvnet}&70.3&80.7&70.3&93.5&44.2&68.4&43.4\\
PVKD\pub{CVPR22}\cite{hou2022point}&71.4&77.5&70.9&92.4&53.5&67.9&50.5\\
\hline
KPConv\pub{ICCV19}\cite{thomas2019kpconv}&58.8&72.7&61.3&90.5&33.4&30.2&11.8\\
KPConv $+~\textbf{\texttt{Ours}}$ &\reshl{61.0}{2.2}&75.0&63.4&91.4&49.0&45.0&36.4\\
      \cdashline{1-8}[1pt/1pt]
SPVNAS$_{\text{10.8M}}$\pub{ECCV20}\cite{tang2020searching}&62.3 & 73.8&63.2&90.9&50.9&40.6&21.8\\
SPVNAS$_{\text{10.8M}}$ +~\textbf{\texttt{Ours}} &\reshl{64.3}{2.0}&73.9&64.0&91.4&48.0&48.9&23.2\\
      \cdashline{1-8}[1pt/1pt]
Cylinder3D\pub{CVPR21}\cite{zhu2021cylindrical} &67.8&75.5&65.1&91.0&50.8&67.6&36.0\\
Cylinder3D +~\textbf{\texttt{Ours}} &\reshl{70.4}{2.6}&77.2&66.1&92.3&51.9&68.4&54.6\\
\hline
\end{tabular}}
\vspace{-13pt}
\end{center}
\end{table}

\noindent\textbf{Quantitative Result.} Table\!~\ref{table:StaticUrbantest} reports comparison results on SemanticKITTI single-scan challenge \texttt{test}. As seen, our algorithm improves
the performance of the base segmentation networks by solid margins. Concretely, it yields \textbf{2.2}\%, \textbf{2.6}\%, and \textbf{2.0}\% mIoU gains over point-based KPConv\!~\cite{thomas2019kpconv},~voxel-based Cylinder3D\!~\cite{zhu2021cylindrical}, and SPVNAS$_{\!}$~\cite{tang2020searching}, respectively. Our algorithm  also obtains consistent~performance improvements across most classes. These results illusrate the wide potential benefit of our algorithm. Moreover, ``Cylinder3D$_{\!}$ +$_{\!}$ \texttt{Ours}'' reaches comparable results with published competitors. This is particularly impressive considering the fact that the improvement is solely achieved by our training scheme, without any network architectural modification and inference speed delay.

\begin{figure*}[t]
\vspace{-6pt}
  \centering
      \includegraphics[width=0.99 \linewidth]{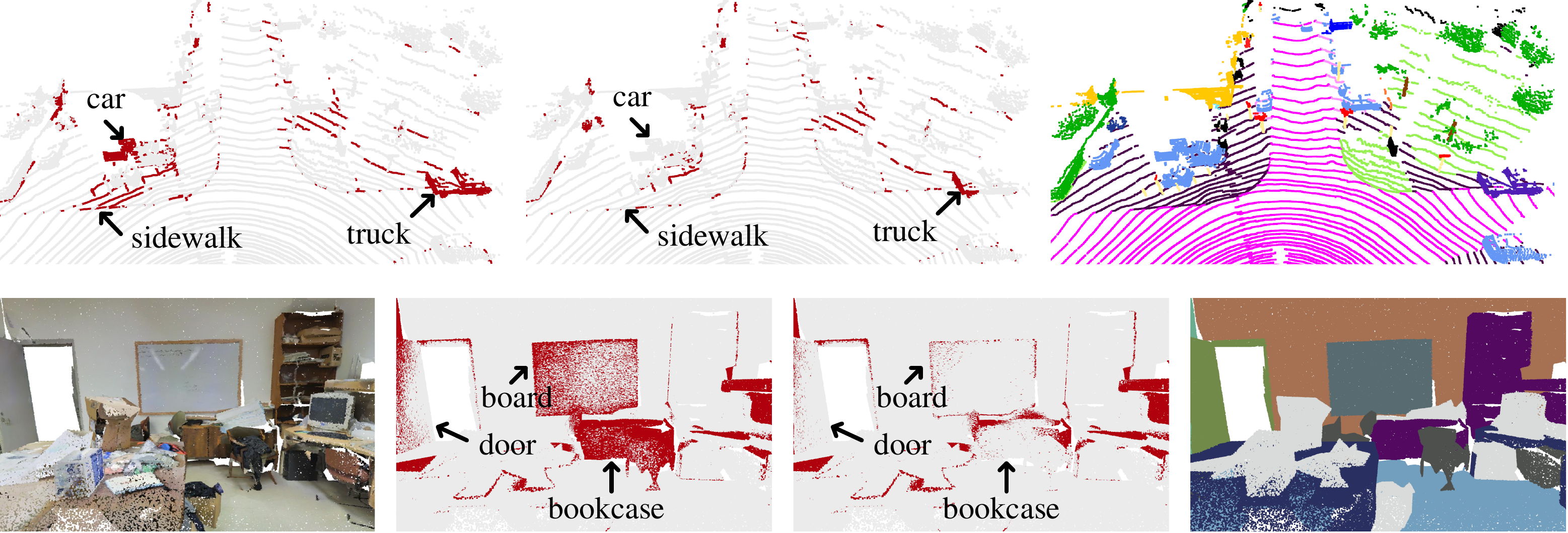}
      \put(-449,76) {\scalebox{.80}{Cylinder3D~\cite{zhu2021cylindrical}}}
      \put(-245,76){\scalebox{.80}{Ours}}
      \put(-95.5,76){\scalebox{.80}{Ground Truth}}
      \put(-439,-7) {\scalebox{.80}{Input}}
      \put(-328,-7) {\scalebox{.80}{PTV1~\cite{zhao2021point}}}
      \put(-188,-7){\scalebox{.80}{Ours}}
      \put(-85,-7){\scalebox{.80}{Ground Truth}}
\vspace{-6pt}
\captionsetup{width=.99\textwidth}
\caption{Error maps on SemanticKITTI~\cite{behley2019semantickitti} single-scan challenge \texttt{val} (\textbf{top}), and S3DIS~\cite{armeni20163d} Area-5 (\textbf{bottom}). The differences are illustrated by arrows.}
\label{fig:vresults1}
\vspace{-15pt}
\end{figure*}
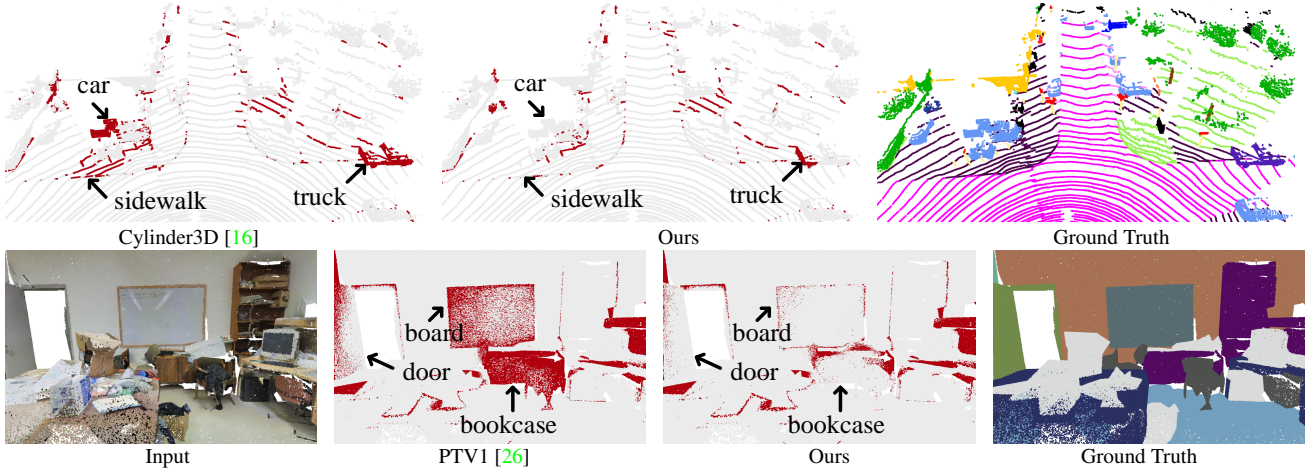

\begin{table}[t]
\begin{center}
\captionsetup{width=.48\textwidth}
\caption{$_{\!}$\textbf{Quantitative$_{\!}$ 3D$_{\!}$ segmentation$_{\!}$ results$_{\!}$} on S3DIS \cite{armeni20163d} Area-5 (\S\!~\ref{sec:ex2}). For clarity, IoUs on 5 of 13 classes are given {\small(c$_1$: \textit{wall}, c$_2$: \textit{column}, c$_3$: \textit{window}, c$_4$: \textit{door}, c$_5$: \textit{board})}.
}
\vspace{-2pt}
\label{table:StaticIndoorval}
\setlength\tabcolsep{2pt}
\renewcommand\arraystretch{1.05}
\resizebox{\linewidth}{!}{
\begin{tabular}{|r||cc|ccccc|}
\thickhline
\rowcolor{mygray}
Method&mIoU(\%)&mAcc(\%)&c$_1$(\%)&c$_2$(\%)&c$_3$(\%)&c$_4$(\%)&c$_5$(\%)\\
\hline
\hline
HPEIN\pub{ICCV19}\cite{jiang2019hierarchical}&61.9&68.3&81.4&23.3&65.3&40.0&65.6\\
PAT\pub{CVPR19}\cite{yang2019modeling}&60.1&70.8&72.3&41.5&85.1&38.2&61.3\\
PointWeb\pub{CVPR19}\cite{zhao2019pointweb}&60.3&66.6&79.4&21.1&59.7&34.8&64.9\\
MinkowskiNet\pub{CVPR19}\cite{choy20194d}&65.4&71.7&86.2&34.1&48.9&62.4&74.4\\
SCF-Net\pub{CVPR21}\cite{fan2021scf}&63.8&-&-&-&-&-&-\\
BAAF-Net\pub{CVPR21}\cite{qiu2021semantic}&65.4&73.1&-&-&-&-&-\\
CGA-Net\pub{CVPR21}\cite{lu2021cga}&68.6&-&83.0&25.3&59.6&71.0&69.5\\
PTV1$+$CBL\pub{CVPR22}\cite{tang2022contrastive}&71.6&77.9&-&-&-&-&-\\
Stratified Trans.\pub{CVPR22}\cite{lai2022stratified}&72.0&78.1&-&-&-&-&-\\
PTV2\pub{NeurIPS22}\cite{wupoint}&72.6&78.0&-&-&-&-&-\\
\hline
KPConv\pub{ICCV19}\cite{thomas2019kpconv}&67.1&72.8&82.4&23.9&58.0&69.0&66.7\\
KPConv$+~\textbf{\texttt{Ours}}$&\reshl{69.0}{1.9}&\reshl{76.2}{3.4}&84.0&30.7&66.7&77.6&63.0\\
\cdashline{1-8}[1pt/1pt]
PTV1\pub{ICCV21}\cite{zhao2021point}&70.4&76.5&86.3&38.0&63.4&74.3&76.0\\
PTV1$+~\textbf{\texttt{Ours}}$&\reshl{\textbf{72.2}}{1.8}&\reshl{\textbf{79.6}}{3.1}&88.1&49.3&65.3&79.4&81.0\\
\hline
\end{tabular}}
\end{center}
\vspace{-20pt}
\end{table}

\noindent\textbf{Qualitative Result.} As shown in the top row of Fig.~\!\ref{fig:vresults1}, our method can reduce errors over both small nature objects (such as \textit{trunk}) and widely distributed classes (like \textit{sidewalk}).

\begin{table}[t]
\vspace{4pt}
\begin{center}
    \captionsetup{width=.48\textwidth}
\caption{$_{\!}$\textbf{Quantitative$_{\!}$ 4D$_{\!}$ segmentation$_{\!}$ results$_{\!}$} on$_{\!}$ SemanticKITTI \cite{behley2019semantickitti}$_{\!}$ multi-scan$_{\!}$ challenge$_{\!}$ \texttt{test}$_{\!}$ (\S\ref{sec:ex3}).$_{\!}$ IoUs on 6 of 25 classes are reported {\small(c$_1$:$_{\!}$ \textit{sidewalk}, c$_2$:$_{\!}$ \textit{moving$_{\!}$ car}, c$_3$:$_{\!}$ \textit{moving$_{\!}$ truck}, c$_4$:$_{\!}$ \textit{bicycle}, c$_5$:$_{\!}$ \textit{motorcyclist}, c$_6$:$_{\!}$ \textit{traffic-sign})}.
}
\vspace{-2pt}
\label{table:4DSegmentationtest1}
\setlength\tabcolsep{3pt}
\renewcommand\arraystretch{1.05}
\resizebox{\linewidth}{!}{
\begin{tabular}{|r||c|cccccc|}
\thickhline
\rowcolor{mygray}
Method&mIoU(\%)&c$_1$(\%)&c$_2$(\%)&c$_3$(\%)&c$_4$(\%)&c$_5$(\%)&c$_{6}$(\%)\\
\hline
\hline
TangentConv\pub{CVPR18}\cite{tatarchenko2018tangent}&34.1&64.0&40.3&1.1&2.0&0.0&31.2\\
DarkNet53\pub{ICCV19}\cite{behley2019semantickitti}&41.6&75.3&61.5&14.1&30.4&0.0&31.2\\
TemporalLidarSeg\pub{3DV20}\cite{duerr2020lidar}&47.0&75.8&68.2&2.1&47.7&0.0&60.4\\
SpSeqnet\pub{CVPR20}\cite{shi2020spsequencenet}&43.1&73.9&53.2&41.2&24.0&0.0&48.7\\
\hline
KPConv\pub{ICCV19}\cite{thomas2019kpconv}&51.2&70.5&69.4&5.8&44.9&0.0&53.9\\
KPConv$+~\textbf{\texttt{Ours}}$&\reshl{53.2}{2.0}&75.2&75.2&4.1&67.2&9.9&64.6\\
\cdashline{1-8}[1pt/1pt]
Cylinder3D\pub{CVPR21}\cite{zhu2021cylindrical}&52.5&74.5&74.9&0.0&67.6&0.2&61.4\\
Cylinder3D$+~\textbf{\texttt{Ours}}$&\reshl{54.7}{2.2}&76.9&81.7&11.9&55.9&3.0&68.0\\
\hline
\end{tabular}}
\end{center}
\vspace{-20pt}
\end{table}

\vspace{-2pt}
\subsection{3D Segmentation on Static Indoor Point Clouds}
\label{sec:ex2}
\vspace{-1pt}
\noindent\textbf{Dataset.} S3DIS$_{\!}$~\cite{armeni20163d} is a famous 3D indoor parsing dataset. It contains 273M points collected from six areas and labeled with 13 classes. Following$_{\!}$~\cite{tchapmi2017segcloud,thomas2019kpconv}, we use Area-5 as test scene to better test the generalization ability. We report two metrics: mIoU and mean of class-wise accuracy (mAcc).

\noindent\textbf{Quantitative Result.} Table~\ref{table:StaticIndoorval} summarizes the comparison results on S3DIS, showing our training algorithm also works well on large-scale challenging indoor point clouds. In particular, our algorithm brings impressive gains over KPConv, \ie, 67.1\%$\rightarrow$\textbf{69.0}\% and 72.8\%$\rightarrow$\textbf{76.2}\%, in terms of mIoU and mAcc. Notably, with PTV1 as the backbone, our approach attains mIoU/mAcc of \textbf{72.2}\%/\textbf{79.6}\%, outperforming PTV1$+$CBL (71.6\%/77.9\%).

\noindent\textbf{Qualitative Result.} As shown in the bottom row of Fig.~\!\ref{fig:vresults1}, our method significantly reduces the errors of PTV1~\cite{zhao2021point} in an indoor environment of S3DIS$_{\!}$~\cite{armeni20163d}$_{\!}$ Area-5.

\subsection{4D Segmentation on Urban Point Sequences}
\label{sec:ex3}
\vspace{-2pt}
\noindent\textbf{Dataset.$_{\!}$} SemanticKITTI$_{\!}$~\cite{behley2019semantickitti}$_{\!}$ multi-scan$_{\!}$ challenge$_{\!}$ is$_{\!}$ devoted$_{\!}$ to$_{\!}$ 4D point cloud segmentation. It involves six more classes to distinguish between moving objects and stationary ones for \textit{car}, \textit{trunk}, \textit{bicyclist}, \textit{other-vehicle}, \textit{person}, and \textit{motor-cyclist} categories. mIoU is adopted as the evaluation metric.

\begin{table}[t]
\vspace{4pt}
\begin{center}
    \captionsetup{width=.48\textwidth}
\caption{$_{\!}$\textbf{Quantitative$_{\!}$ 3D$_{\!}$ detection$_{\!}$ results$_{\!}$} on KITTI~\cite{geiger2013vision} challenge \texttt{val} (\S\ref{sec:detection}).}
\label{table:kittidetection}
\vspace{-2pt}
\setlength\tabcolsep{4pt}
\renewcommand\arraystretch{1.1}
\resizebox{\linewidth}{!}{
\begin{tabular}{|c|r||c|ccc|}
\thickhline
\rowcolor{mygray}
\hline Difficulty & Method & mAP(\%) & Car(\%) & Pedestrian(\%) & Cyclist(\%) \\
\hline \hline
\multirow{4}{*}{Easy} & Second\pub{SENSORS18} \cite{yan2018second} & 75.25 & 88.61 & 56.55 & 80.59 \\
& Second +~\texttt{\textbf{Ours}} & \reshl{78.60}{3.35} & 89.13 & 58.50 & 88.16 \\
\cdashline{2-6}[1pt/1pt]
& PointPillar\pub{CVPR19} \cite{Lang_2019_CVPR} & 74.76 & 86.46 & 57.75 & 80.06 \\
& PointPillar +~\texttt{\textbf{Ours}} & \reshl{76.82}{2.06} & 88.34 & 58.19 & 83.92 \\
\hline
\multirow{4}{*}{Moderate} & Second\pub{SENSORS18} \cite{yan2018second} & 66.25 & 78.62 & 52.98 & 67.16 \\
& Second +~\texttt{\textbf{Ours}} & \reshl{69.67}{3.42} & 82.97 & 55.64 & 70.39 \\
\cdashline{2-6}[1pt/1pt]
& PointPillar\pub{CVPR19} \cite{Lang_2019_CVPR} & 64.08 & 77.28 & 52.29 & 62.68 \\
& PointPillar +~\texttt{\textbf{Ours}} & \reshl{66.07}{1.99} & 78.43 & 53.31 & 66.47 \\
\hline
\multirow{4}{*}{Hard} & Second\pub{SENSORS18} \cite{yan2018second} & 62.69 & 77.22 & 47.73 & 63.11 \\
& Second +~\texttt{\textbf{Ours}} & \reshl{65.36}{2.67} & 78.55 & 50.91 & 66.61 \\
\cdashline{2-6}[1pt/1pt]
& PointPillar\pub{CVPR19} \cite{Lang_2019_CVPR} & 60.76 & 74.65 & 47.91 & 59.71 \\
& PointPillar +~\texttt{\textbf{Ours}} & \reshl{62.96}{2.20} & 77.14 & 49.15 & 62.61 \\
\hline
\end{tabular}}
\end{center}
\vspace{-20pt}
\end{table}

\begin{table*}
\begin{center}
\vspace{-2pt}
    \captionsetup{width=.98\textwidth}
\caption{${\!}$\textbf{Study$_{\!}$ of$_{\!}$ proposed$_{\!}$ training$_{\!}$ strategy}$_{\!}$ on$_{\!}$ S3DIS~\cite{armeni20163d}$_{\!}$ Area-5$_{\!}$ and$_{\!}$ SemanticKITTI~\cite{behley2019semantickitti}$_{\!}$ multi-scan$_{\!}$ \texttt{val}$_{\!}$ set$_{\!}$$_{\!}$ (\S\ref{sec:ablation}).}
\vspace{-2pt}
\label{table:AblationStudies}
\setlength\tabcolsep{8pt}
\renewcommand\arraystretch{1.0}
\resizebox{0.95\linewidth}{!}{
\begin{tabular}{|r||cc|cc|c|}
\thickhline
\rowcolor{mygray}
 &$\mathcal{J}_{\text{PPC}}${{ (Eq.\!~(\ref{eq:pNCE}))}} &$\mathcal{J}_{\text{PCC}}${{ (Eq.\!~(\ref{eq:pNCE2}))}} &{S3DIS}{\scriptsize{ mIoU(\%)}} &{S-KITTI}{\scriptsize{ mIoU(\%)}} &Training Speed{\scriptsize{ (sec/epoch)}}\\
\hline
\hline
Baseline (\textit{w/o} clustering analysis) && &67.1 &53.3 &281.46\\
\hline
Point-Point Contrast  &\checkmark&  &68.0 & 54.4 &310.20\\
Point-Center Contrast &&\checkmark &68.4 &54.7 &310.28\\
Point-Point$_{\!}$ +$_{\!}$ Point-Center Contrast &\checkmark &\checkmark &\textbf{69.0} &\textbf{55.7} &311.71\\
\hline
\end{tabular}
}
\vspace{-13pt}
\end{center}
\end{table*}

\begin{figure*}[t]
\begin{minipage}{\textwidth}
    \hspace{3pt}
    \begin{minipage}[t]{0.32\textwidth}
    \captionsetup{width=1.\textwidth}
     \makeatletter\def\@captype{table}\makeatother\caption{\textbf{$_{\!}$Curve$_{\!}$ of$_{\!}$ CE$_{\!}$ Loss}$_{\!}$  on$_{\!}$ Semantic- KITTI$_{\!}$~\cite{behley2019semantickitti}$_{\!}$ single-scan$_{\!}$ challenge$_{\!}$ \texttt{train}$_{\!}$ (\textbf{left})$_{\!}$ and$_{\!}$ \texttt{val}$_{\!}$ (\textbf{right}).$_{\!\!\!\!\!}$}
        \vspace{-9pt}
      \includegraphics[width=0.99 \linewidth]{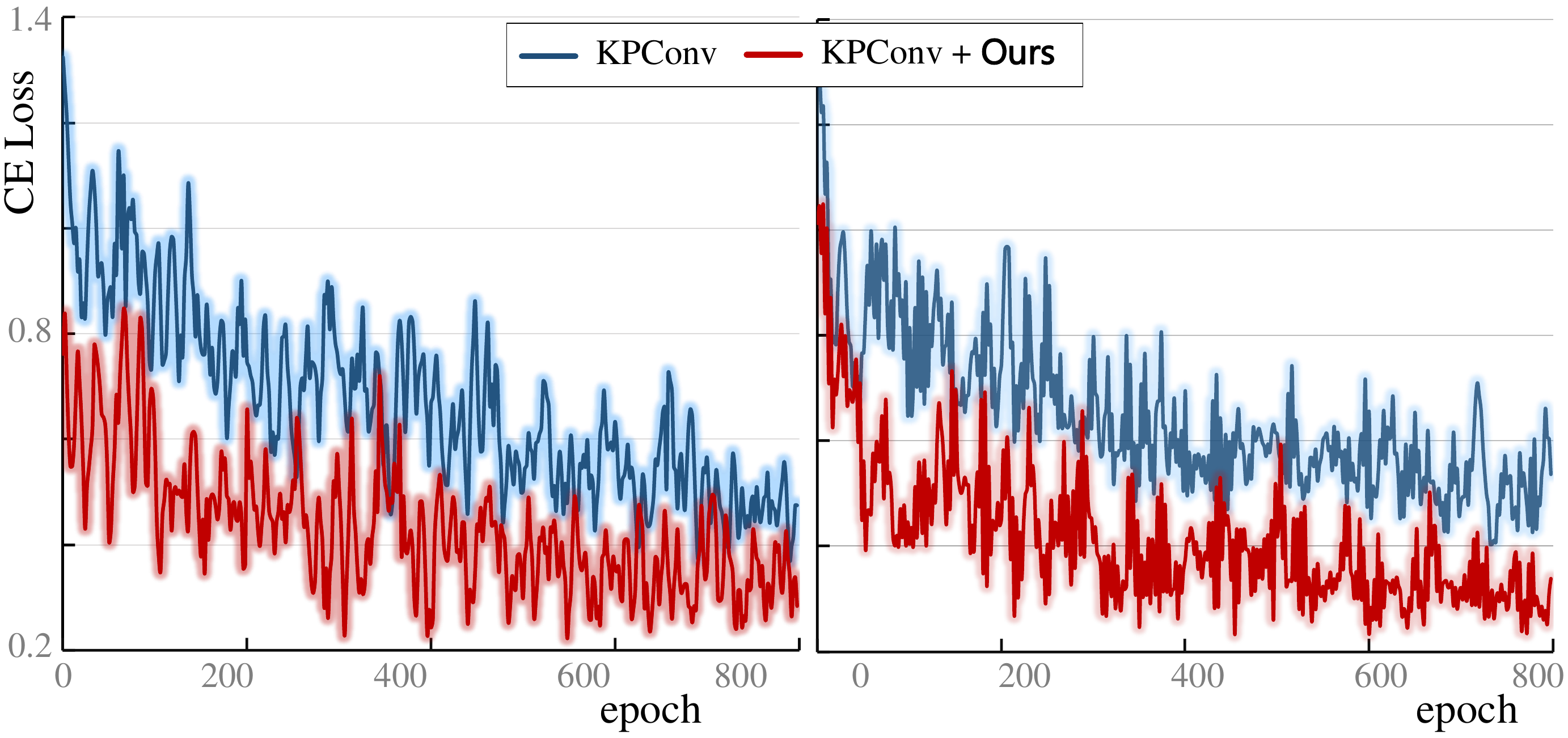}
    \end{minipage}\hspace{6pt}
    \begin{minipage}[t]{0.63\textwidth}
    \captionsetup{width=.95\textwidth}
     \makeatletter\def\@captype{table}\makeatother\caption{$_{\!}$\textbf{$_{\!}$Parameter$_{\!}$ studies$_{\!}$} on$_{\!}$ S3DIS$_{\!}$~\cite{armeni20163d}$_{\!}$ Area-5$_{\!}$ and$_{\!}$ SemanticKITTI$_{\!}$~\cite{behley2019semantickitti}$_{\!}$ (S-KITTI)$_{\!}$ multi-scan$_{\!}$ \texttt{val}$_{\!}$ set$_{\!}$ (\S\ref{sec:ablation}).$_{\!}$ (mIoU(\%)$_{\!}$ is$_{\!}$ reported.)}
  \vspace{-6pt}
  \label{tab:ablations}
  \hspace{-.3em}
    \resizebox{0.302\textwidth}{!}{
    \setlength\tabcolsep{2pt}
\renewcommand\arraystretch{1.01}
    \begin{tabular}{|l||cc|}\thickhline
      \rowcolor{mygray}
      \# Cluster & {S3DIS} &{S-KITTI} \\ \hline\hline
      ~$M=1$  & 67.5 &53.7 \\
      ~$M=10$  & 68.0 &54.5 \\
      ~$M=20$  & 68.5 &55.2 \\
      ~$M=\mathbf{40}\!\!\!$ & 69.0 &55.7\\
      ~$M=60$  & 68.9 &55.2 \\
      ~$M=80$  & 68.7 &55.5\\
      \hline
    \end{tabular}
    \label{table:prototypenumber}
  }\hspace{-0.7em}
    \setlength\tabcolsep{2pt}
\renewcommand\arraystretch{1.0}
    \resizebox{0.376\textwidth}{!}{
    \begin{tabular}{|c||cc|}\thickhline
      \rowcolor{mygray}
      Memory Capacity & {S3DIS} &{S-KITTI} \\ \hline\hline
      \tabincell{c}{Mini-Batch\\(\textit{w/o} memory)} & 68.0 &54.4\\
      ~5 $\times$ \#scene & 68.6 &55.0 \\
            \textbf{10 $\times$ \#scene} & 69.0 & 55.7 \\
            15 $\times$ \#scene & 68.8 &55.7 \\
            20 $\times$ \#scene & 68.7 &55.6 \\
      \hline
    \end{tabular}
    \label{table:capacity}
  }\hspace{-0.7em}
    \resizebox{0.334\textwidth}{!}{
    \setlength\tabcolsep{2pt}
\renewcommand\arraystretch{1.02}
    \begin{tabular}{|l||cc|}\thickhline
      \rowcolor{mygray}
      Coefficient $\mu$ & {S3DIS} &{S-KITTI} \\\hline\hline
      $\mu=0$ &67.7 & 53.6 \\
      $\mu=0.9$ & 68.0 &54.0 \\
      $\mu=0.99$ & 68.5 &54.7 \\
      $\mu=0.999$ & 68.6 &55.3 \\
      $\mathbf{\mu=0.9999}$ & 69.0 & 55.7 \\
            $\mu=0.99999$ &68.8 &55.5\\
      \hline
    \end{tabular}
    \label{table:momentum}
  }
  \put(-303,-46) {{\small (a) $_{\!}$Per-class$_{\!}$ cluster$_{\!}$ Num$_{\!}$}}
  \put(-201,-46){{\small (b) Per-cluster memory}}
  \put(-98,-46){{\small (c) Momentum coefficient}}
    \end{minipage}
  \end{minipage}
  \vspace{-12pt}
\end{figure*}

\noindent\textbf{Quantitative Result.} Table~\ref{table:4DSegmentationtest1} reports our comparison results on SemanticKITTI~\cite{behley2019semantickitti} multi-scan challenge \texttt{test}. Our algorithm, again,  leads to improvements over backbones, \ie, \textbf{2.0}\%  and \textbf{2.2}\%  mIoU gain compared with KPConv~\cite{thomas2019kpconv} and Cylinder3D~\cite{zhu2021cylindrical}, respectively. This confirms our algorithm is also applicable in point cloud sequences. Our algorithm also obtains superior performance for vehicle categories with moving patterns, such as \textit{moving car}, \textit{moving truck}, \textit{moving other-vehicle}, \textit{etc}. We attribute this to our capacity of capturing complex patterns and variations, which improves the robustness in dynamic scenes.

\vspace{-3pt}
\subsection{3D Detection on Static Urban Point Clouds}
\label{sec:detection}
\vspace{-1pt}
To fully reveal the power of our idea, we conduct additional experiments on 3D object detection.

\noindent\textbf{Algorithmic Modification.} To apply our algorithm to the 3D object detection task and minimize the modification effort, we view the bounding box annotations as a form of coarse segmentation labels. For each labeled bounding box with semantic class $c\!\in\!\mathcal{C}$, we simply treat all the points within the bounding box as data examples of class $c$, which are used in our clustering analysis based representation learning (\textit{cf}.~Eqs.\!~\ref{eq:pNCE}-\ref{eq:pNCE2}). Note that there is no change to the base 3D detection network, including the detection head.

\noindent\textbf{Dataset.} KITTI \cite{geiger2013vision} is a standard benchmark for 3D object detection. We split 3712 scans for \texttt{train} and 3769 scans for \texttt{val}, with 3D bounding box annotations of vehicles, pedestrians and cyclists. Detection outcomes~are evaluated under three regimes: \textit{easy, moderate, hard}, defined according to occlusion and truncation levels of
objects. Average precisions are reported with IoU thresholds of 0.7, 0.5, and 0.5, respectively for \textit{car}, \textit{pedestrian}, and \textit{cyclist} classes.

\noindent\textbf{Base Detection Networks.} We apply our algorithm to two famous$_{\!}$ 3D$_{\!}$ detectors,$_{\!}$ \ie,$_{\!}$ Second$_{\!}$~\cite{yan2018second}$_{\!}$ and$_{\!}$ PointPillar$_{\!}$~\cite{lang2019pointpillars}.

\noindent\textbf{Quantitative Result.} Table\!~\ref{table:kittidetection} reports the experimental results on KITTI~\texttt{val}. We can observe that, for both Second and PointPillar, our training algorithm brings notable performance gains, across different classes and under different regimes. This proves the high versatility of our algorithm.

\subsection{Diagnostic Experiment}
\label{sec:ablation}
\vspace{-1pt}
To$_{\!}$  test$_{\!}$  the$_{\!}$  efficacy$_{\!}$  of$_{\!}$  our$_{\!}$  core$_{\!}$  algorithm$_{\!}$  designs,$_{\!}$  we$_{\!}$  con- duct
a series of ablative studies on S3DIS${\!}$~\cite{armeni20163d} Area-5 and$_{\!}$ SemanticKITTI${\!}$~\cite{behley2019semantickitti}$_{\!}$ multi-scan$_{\!}$ challenge$_{\!}$ \texttt{val}.$_{\!}$ We$_{\!}$ adopt$_{\!}$ KPConv${\!}$~\cite{thomas2019kpconv}$_{\!}$  as$_{\!}$ our$_{\!}$ base$_{\!}$ segmentation$_{\!}$ network.$_{\!}$ The$_{\!}$ results$_{\!}$ are$_{\!}$ reported$_{\!}$ without$_{\!}$ post-processing$_{\!}$ or$_{\!}$ test-time$_{\!}$ augmentation.

\noindent\textbf{Clustering$_{\!}$ Analysis$_{\!}$ based$_{\!}$ Network$_{\!}$ Training.$_{\!}$} We$_{\!}$ first$_{\!}$~test the efficacy of our core idea of clustering analysis based point representation learning. As shown in Table\!~\ref{table:AblationStudies}, the baseline model, trained in the standard strategy, gains 67.1\% and 53.3\% mIoU, on S3DIS and SemanticKITTI, respectively. Additionally considering point-point contrast $\mathcal{J}_{\text{PPC}\!}$ (Eq.\!~(\ref{eq:pNCE})) or point-center contrast $\mathcal{J}_{\text{PCC}\!}$ (Eq.\!~(\ref{eq:pNCE2})) can lead to better performance. However, combining these two training objectives yields the best results, \ie, 69.0\% and 55.7\%. These results verify that mining latent data structures can$_{\!}$ benefit$_{\!}$ detailed$_{\!}$ analysis$_{\!}$ of$_{\!}$ point$_{\!}$ cloud.$_{\!}$ Table$_{\!}$~\ref{table:AblationStudies}$_{\!}$ also$_{\!}$ gives comparisons for training speed. Our algorithm only brings negligible delay ($\sim$30 s for each epoch), confirming its high efficiency.

\noindent\textbf{Per-Class Cluster Number $M$.} We next investigate the impact of the cluster number $M$ of each class. The results are summarized in Table\!~\ref{table:prototypenumber}a.  $M\!=\!1$ means that directly treating each class as a single cluster. This baseline obtains 67.5\% and 53.7\% mIoU, on S3DIS and SemanticKITTI, respectively. After clustering based fine-grained pattern mining, we observe consistent improvements, \eg, 67.5\%$\rightarrow$69.0\% on S3DIS when $M\!=\!40$. This verifies that \textbf{i)} there indeed exist some latent patterns in point clouds, and \textbf{ii)} these latent patterns are valuable for point cloud parsing. When $M\!>\!40$, further increasing $M$ gives marginal performance gains even worse results. We speculate this is because the model is distracted by some trivial patterns due to over-clustering.

\noindent\textbf{Memory Bank.} Then we study the influence of our memory bank in Table~\ref{table:capacity}b. ``Mini-Batch (\textit{w/o} memory)'' means that only computing contrast within each mini-batch, without the memory; it earns 68.0\% and 54.4\% mIoU, on S3DIS and SemanticKITTI, respectively. We then provision this baseline with class-wise memory bank with different capacities. When storing 10 point features per scene for each cluster, the best performance is achieved, \ie, 69.0\% and 55.7\%.

\noindent\textbf{Momentum${\!}$ Coefficient${\!}$ $\mu$.${\!}$} Table${\!}$~\ref{table:momentum}c${\!}$ gives${\!}$ the${\!}$ performance with${\!}$~regard${\!}$ to${\!}$ the${\!}$ momentum${\!}$ coefficient${\!}$ $\mu$${\!}$ (\textit{cf}.${\!}$~Eq.\!~\ref{eq:update}),${\!}$ which controls the evolution speed of cluster centers. The model performs better with a relatively large coefficient (\ie, $\mu\!=\!0.9999$), showing that slow update is more favored. Moreover, at the extreme case of $\mu\!=\!0$, the performance drops considerably, evidencing that simply approximating the cluster centers with per-batch cluster means is not a sound solution.

\section{Conclusion and Discussion}
We devise a clustering based supervised training scheme for point cloud analysis, which discovers and respects latent data structures during point representation learning. Rather than simply minimizing~the point recognition error, we iteratively$_{\!}$ perform$_{\!}$ 1)$_{\!}$ unsupervised,$_{\!}$ within-class$_{\!}$ clustering$_{\!}$ based subclass pattern$_{\!}$ mining,$_{\!}$ and$_{\!}$ 2)$_{\!}$ clustering$_{\!}$ assignment$_{\!}$ based$_{\!}$ point$_{\!}$ embedding$_{\!}$ space optimization. Our algorithm~is general and shows outstanding performance over various tasks and datasets. It also brings some new challenges, inc- luding the extension in instance-aware$_{\!}$ segmentation$_{\!}$ setting,$_{\!}$ and$_{\!}$ automatic$_{\!}$ estimation$_{\!}$ of$_{\!}$ the$_{\!}$ cluster$_{\!}$ number.

\clearpage
\renewcommand{\thefigure}{S\arabic{figure}}
\renewcommand{\thetable}{S\arabic{table}}
\renewcommand{\thealgorithm}{S\arabic{algorithm}}
\setcounter{figure}{0}
\setcounter{table}{0}

\appendix

In this supplementary material, we provide the following sections for a better understanding of the main paper. The pseudo-code of clustering based point cloud segmentation learning is elaborated in \S\ref{appendix1}. \S\ref{appendix3} presents the distribution of point data over cluster centers. More qualitative and quantitative results are further presented and analyzed in \S\ref{appendix4}. Finally, limitation and societal impact are discussed in \S\ref{appendix5}.

\section{Pseudo-Code}
\label{appendix1}
Algorithm~\ref{alg:algorithm1} provides a pseudo-code of `assigning subclass labels' function and `update operation' function. Correspondingly, Algorithm~\ref{alg:algorithm2} provides a pseudo-code of $\mathcal{J}_{\text{PCC}}$ (see Eq.~(\ref{eq:pNCE2})). The implementation of $\mathcal{J}_{\text{PPC}}$ is similar to it, so we do not show pseudo-code for $\mathcal{J}_{\text{PPC}}$. Moreover, to guarantee the reproducibility, our code is released at: \url{github.com/FengZicai/Cluster3Dseg}.


\section{Distribution of Subclass Clusters}
\label{appendix3}
\begin{figure*}[t]
  \centering
      \includegraphics[width=0.99\linewidth]{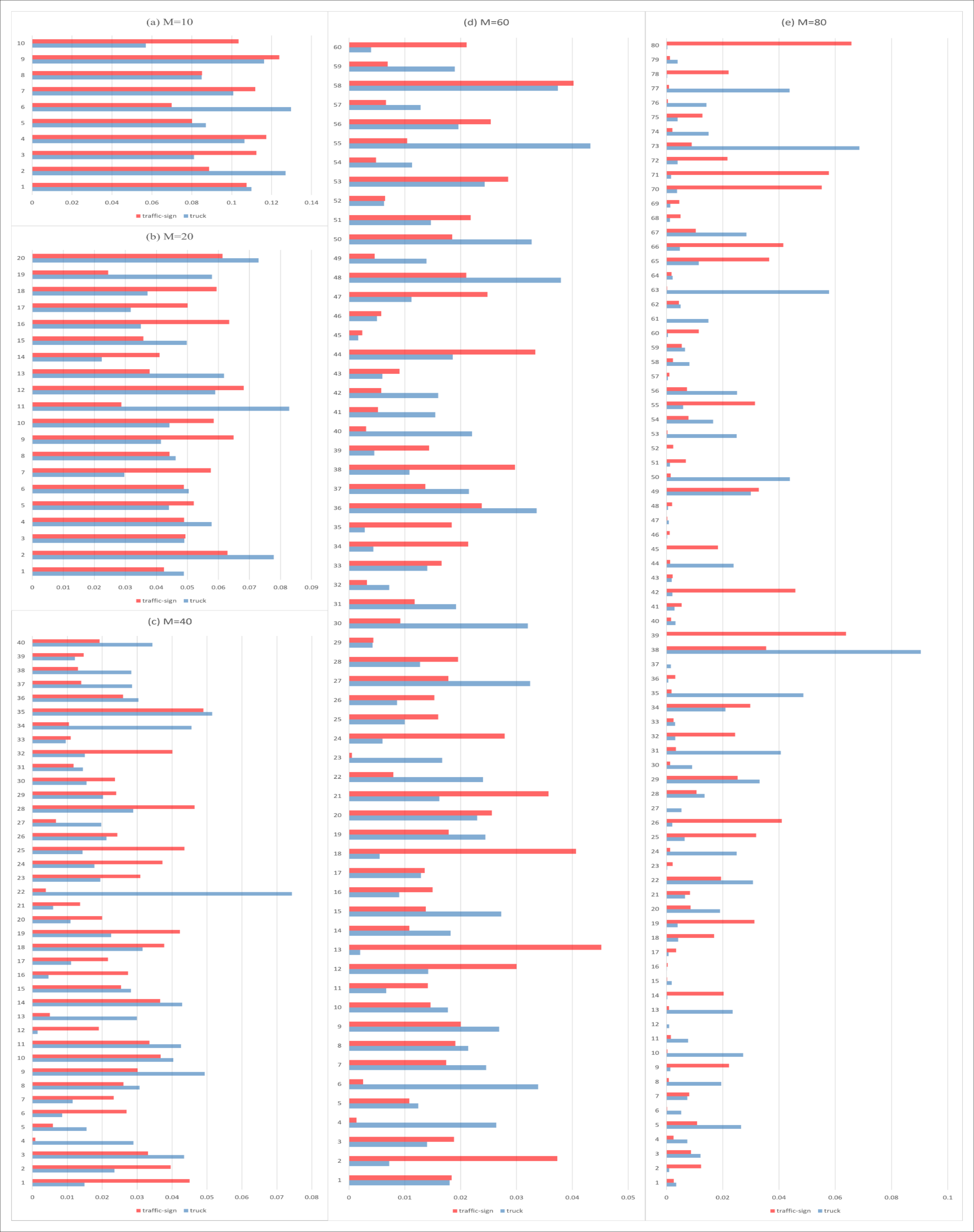}
\captionsetup{width=.99\textwidth}
\caption{Distribution plot with different numbers $M=\{10, 20, 40, 60, 80\}$ of clusters for `truck' and `traffic-sign' classes. (Best viewed with zoom-in.)}
\vspace{-8pt}
\label{fig:dis}
\end{figure*}

Fig.~\!\ref{fig:dis} shows point assignment distribution for `truck' and `traffic-sign' classes, with different numbers $M=\{10, 20, 40, 60, 80\}$ of clusters. We can find that 1) the number of point samples assigned to each cluster is different; 2) with the increase of $M$, some sub-class centers only contain a limited number of samples, especially when $M=80$. In this case, the value of $M$ has exceeded the number of underlying subclass centers in the dataset, resulting in over-clustering. And therefore, some trivial patterns may distract the model and cause performance degradation.

\section{More Qualitative and Quantitative Results}
\label{appendix4}

\noindent\textbf{Complete Quantitative Result on SemanticKITTI Single-Scan Challenge \texttt{test}.} Table~\ref{table:StaticUrbantestappendix} and \ref{table:StaticUrbantestappendix2} report the complete results on SemanticKITTI$_{\!}$~\cite{behley2019semantickitti}$_{\!}$ single-scan$_{\!}$ challenge$_{\!}$ \texttt{test}. Our method reaches 70.4\% mIoU, which yields 2.6\% mIoU gains over Cylinder3D\!~\cite{zhu2021cylindrical}. Moreover, it also outperforms many famous segmentation models, such as AF2S3Net\!~\cite{cheng20212} and RPVNet\!~\cite{xu2021rpvnet}. One more thing to point out, spvnas\footnote{\url{https://github.com/mit-han-lab/spvnas/}} did not provide the source code of 3D-NAS pipeline and the control file for SPVNAS$_{\text{12.5M}}$. But the control file and pretrained models for SPVNAS$_{\text{10.8M}}$ are shared\footnote{SPVNAS has cancelled the download link for the Control file and SPVNAS$_{\text{10.8M}}$ model. Instead, we will release the two previously downloaded files.}. And the difference between SPVNAS$_{\text{12.5M}}$ and SPVNAS$_{\text{10.8M}}$ is that SPVNAS$_{\text{10.8M}}$ is trained except sequence 08. As for our implementation, SPVNAS$_{\text{10.8M}}$ and SPVNAS$_{\text{10.8M}}$ +~\textbf{\texttt{Ours}} are trained on sequences 00-10 and evaluated on 11-21.

\noindent\textbf{Complete Quantitative Result on S3DIS Area-5.} Table~\ref{table:StaticIndoorvalappendix} and \ref{table:StaticIndoorvalappendix2} present the complete per-class IoU on S3DIS~\cite{armeni20163d} Area-5. Both CBL\cite{tang2022contrastive} and our method use contrastive loss on the premise of fully supervised learning. But \cite{tang2022contrastive} only samples negative points locally around the boundaries, while we contrast global subclass centers against the points sampled from the ENTIRE training dataset. Our idea is much more powerful and insightful. The fair comparison based on PTV1\cite{zhao2021point} shows that our approach attains mIoU/mAcc of 72.2\%/79.6\%, outperforming PTV1+CBL (71.6\%/77.9\%).

\noindent\textbf{Complete Quantitative Result on SemanticKITTI multi-scan challenge \texttt{test}.} Table~\ref{table:4DSegmentationtest1appendix} and \ref{table:4DSegmentationtest1appendix2} report the complete results on SemanticKITTI~\cite{behley2019semantickitti} multi-scan challenge \texttt{test}. With Cylinder3D, our algorithm also attains consistent performance improvements of 2.2\% mIoU, just like that in single-scan \texttt{test}. Moreover, Cylinder3D$+~\textbf{\texttt{Ours}}$ surpasses Cylinder3D in 17 classes out of 25 classes.

\begin{figure*}[h!]
  \centering
      \includegraphics[width=0.98 \linewidth]{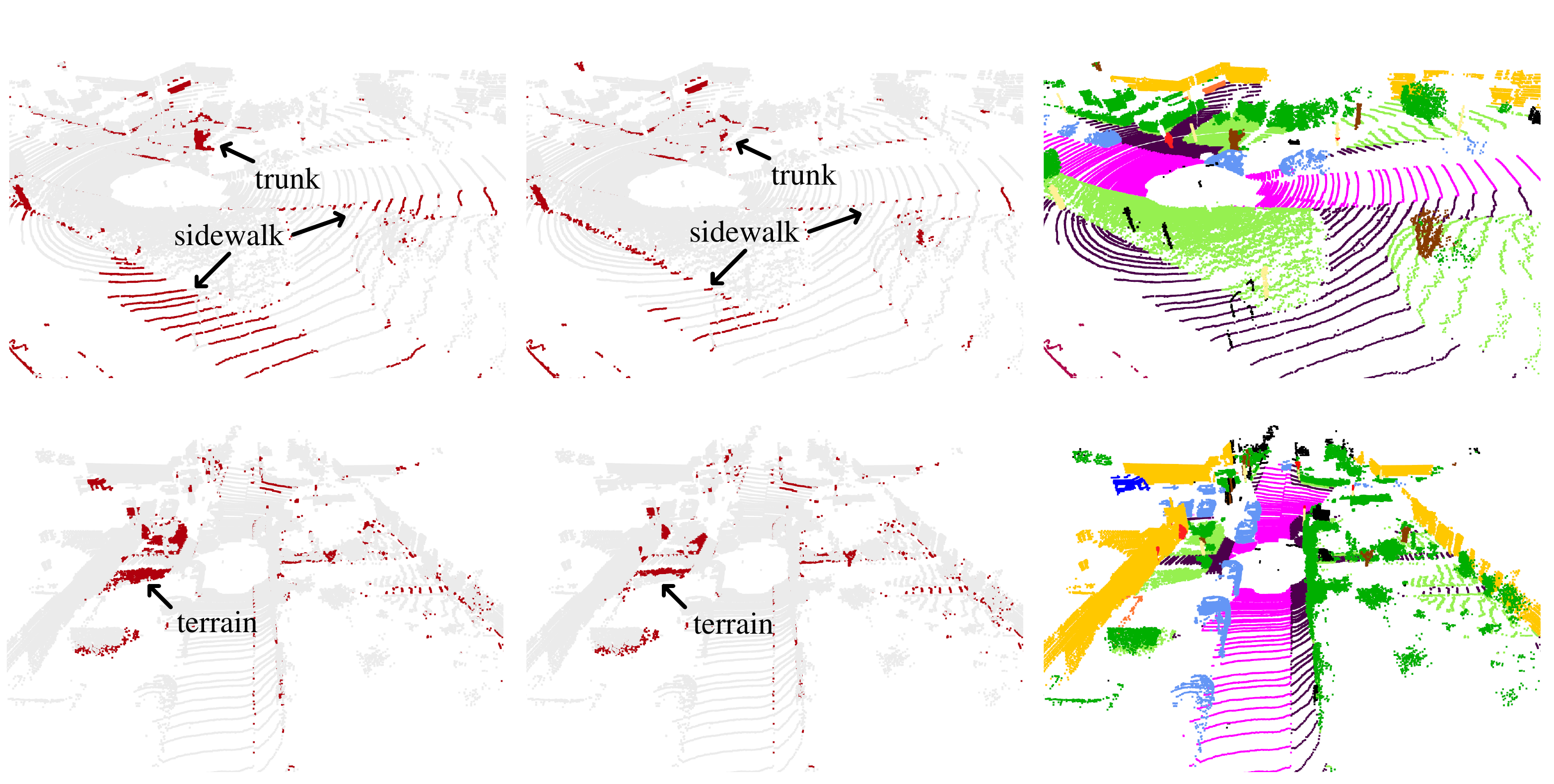}
      \put(-430.5,-6) {\scalebox{.80}{Cylinder3D~\cite{zhu2021cylindrical}}}
      \put(-255.5,-6){\scalebox{.80}{Ours}}
      \put(-100.5,-6){\scalebox{.80}{Ground Truth}}
\captionsetup{width=.99\textwidth}
\caption{Error maps of Cylinder3D \cite{zhu2021cylindrical} and Ours on SemanticKITTI~\cite{behley2019semantickitti} single-scan challenge \texttt{val}$_{\!}$ (\S\ref{sec:ex1}). The differences are as illustrated by arrows.}
\label{fig:vresults1appendix}
\end{figure*}

\begin{figure*}[h!]
  \centering
      \includegraphics[width=0.98 \linewidth]{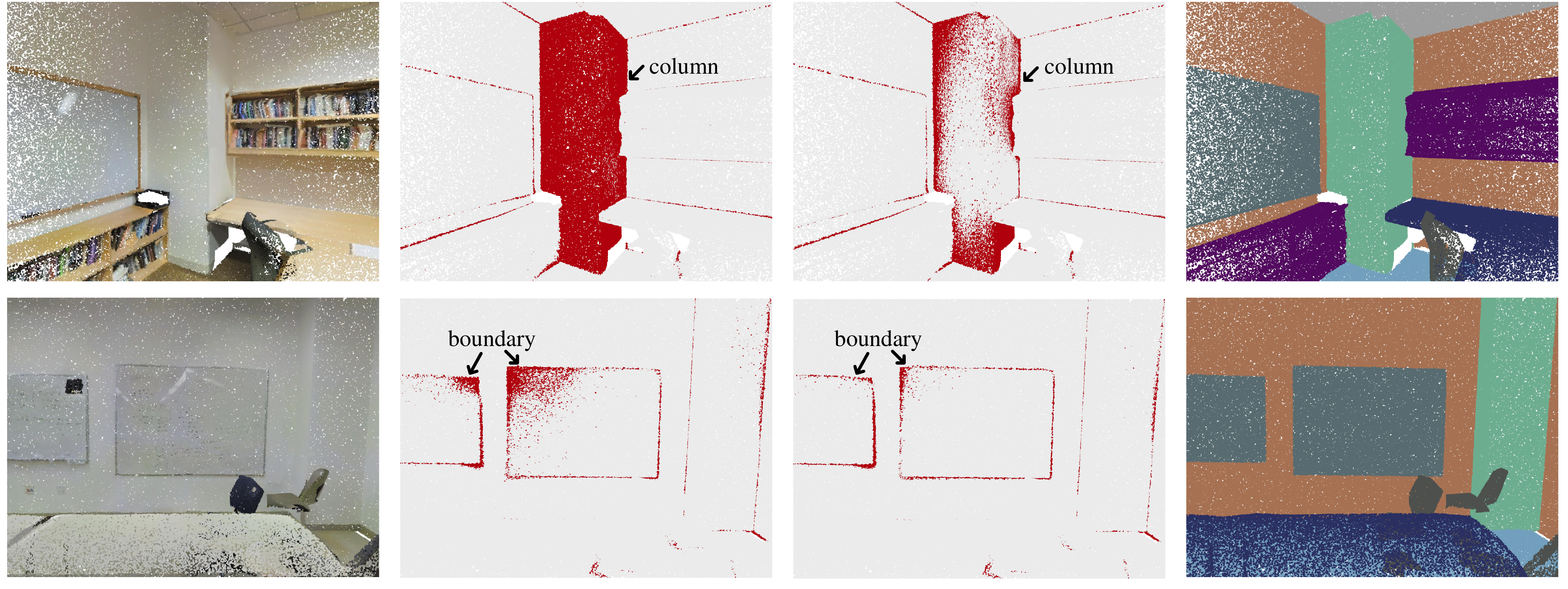}
      \put(-440,-6) {\scalebox{.80}{Input}}
      \put(-325.5,-6) {\scalebox{.80}{PTV1~\cite{zhao2021point}}}
      \put(-190.5,-6){\scalebox{.80}{Ours}}
      \put(-91,-6){\scalebox{.80}{Ground Truth}}
\captionsetup{width=.99\textwidth}
\caption{Error maps of PTV1 \cite{zhao2021point} and Ours on S3DIS~\cite{armeni20163d} Area-5$_{\!}$ (\S\ref{sec:ex2}). The differences are as illustrated by arrows.}
\label{fig:vresults2appendix}
\end{figure*}

\begin{figure*}[h!]
  \centering
      \includegraphics[width=0.98 \linewidth]{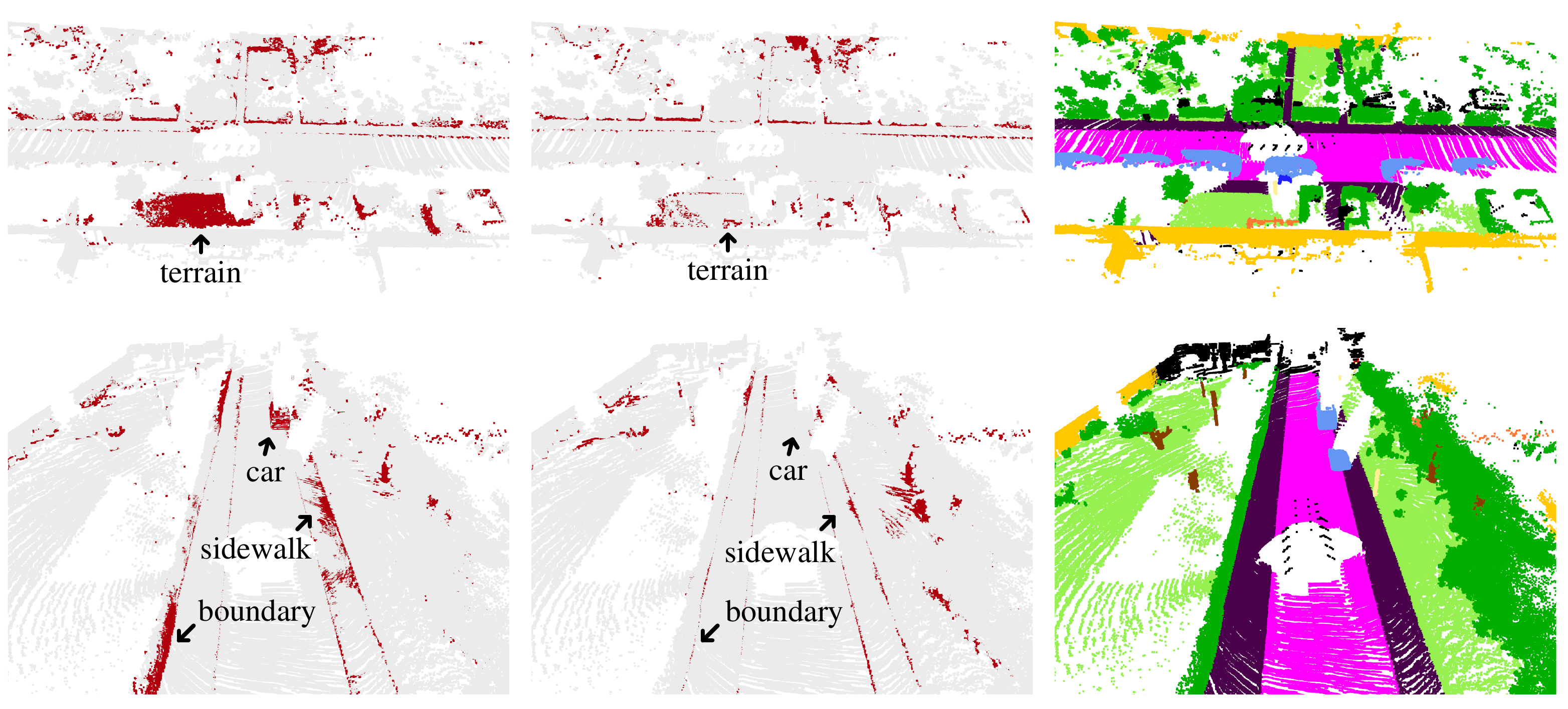}
      \put(-430.5,-6) {\scalebox{.80}{Cylinder3D~\cite{zhu2021cylindrical}}}
      \put(-255.5,-6){\scalebox{.80}{Ours}}
      \put(-100.5,-6){\scalebox{.80}{Ground Truth}}
\captionsetup{width=.99\textwidth}
\caption{Error maps of Cylinder3D \cite{zhu2021cylindrical} and Ours on SemanticKITTI~\cite{behley2019semantickitti} multi-scan challenge \texttt{val}$_{\!}$ (\S\ref{sec:ex3}). The differences are as illustrated by arrows.}
\label{fig:vresults3appendix}
\end{figure*}

\noindent\textbf{Qualitative Results for Segmetation.} We show more qualitative results on SemanticKITTI$_{\!}$~\cite{behley2019semantickitti}$_{\!}$ single-scan$_{\!}$ challenge$_{\!}$ \texttt{val}$_{\!}$ (Fig.~\!\ref{fig:vresults1appendix}), S3DIS$_{\!}$~\cite{armeni20163d}$_{\!}$ Area-5$_{\!}$ (Fig.~\!\ref{fig:vresults2appendix}) and SemanticKITTI$_{\!}$~\cite{behley2019semantickitti}$_{\!}$ multi-scan$_{\!}$ challenge$_{\!}$ \texttt{val}$_{\!}$ (Fig.~\!\ref{fig:vresults3appendix}). As observed, our approach generally gives more accurate predictions compared with vanilla PTV1\cite{zhao2021point} and Cylinder3D\cite{zhu2021cylindrical}. In Fig.~\!\ref{fig:vresults2appendix}, vanilla PTV1 fails to recognize region boundaries and tends to misclassify board-like objects, while our method can significantly reduce these errors. Fig. \!\ref{fig:vresults3appendix} depicts qualitative comparisons of Cylinder3D and Cylinder3D +~\texttt{\textbf{Ours}} over lidar sequences on SemanticKITTI multi-scan challenge \texttt{val}. Note that, the predicted labels of five consecutive frames are displayed in one frame. It can be observed that Cylinder3D~+~\texttt{\textbf{Ours}} has smaller errors over the semantic boundaries as well as classes belonging to ground and nature.

\section{Limitation and Societal Impact}
\label{appendix5}
\noindent\textbf{License of Assets.} Cylinder3D\footnote{\url{https://github.com/xinge008/Cylinder3D/}} is released with Apache license. KPConv\footnote{\url{https://github.com/HuguesTHOMAS/KPConv-PyTorch/}} is implemented based on its released code with MIT license. We have also implemented our method on Point Transformer\footnote{\url{https://github.com/POSTECH-CVLab/point-transformer}}. SPVNAS\footnote{\url{https://github.com/mit-han-lab/spvnas/}} is implemented based on its released code with MIT license. For 3D object detection, our implementation is based on OpenPCDet\footnote{\url{https://github.com/HuguesTHOMAS/KPConv-PyTorch/}}, and it is released under the Apache 2.0 license. Our code is released at: \url{github.com/FengZicai/Cluster3Dseg}.

\noindent\textbf{Limitation.} For some very rare classes, such as \textit{beam} in S3DIS~\cite{armeni20163d}, \textit{bicyclist} in SemanticKITTI~\cite{behley2019semantickitti} multi-scan challenge, our algorithm did not show better-improved results. However, many previous state-of-the-arts~\cite{thomas2019kpconv,zhao2021point,zhu2021cylindrical} also perform poorly on these classes. In the future, we plan to explore smarter data sampling strategies and hard example synthesis techniques to address this issue.

\noindent\textbf{Societal Impact.} For the potential negative societal impacts, in real-world robot navigation tasks or autonomous driving tasks, inaccurate prediction of point cloud labels may lead agents to the wrong category and raise human safety concerns. To avoid this potential problem, we suggest proposing a security protocol in case of dysfunction of our algorithm in real-world applications.

\clearpage

\begin{algorithm*}[h!]
\caption{Pseudo-code of clustering based point cloud segmentation learning - Part I.}
\label{alg:algorithm1}
\begin{algorithmic}[0] 
\STATE    $\color{agreen}\texttt{\# M: number of subclusters.}$
\STATE    $\color{agreen}\verb|# nc: number of classes.|$
\STATE    $\color{agreen}\verb|# dim: number of dimensions.|$
\STATE    $\color{agreen}\verb|# x: features (N, dim).|$
\STATE    $\color{agreen}\verb|# | \hat{\texttt{y}} \verb|: labels (N).|$
\STATE    $\color{agreen}\verb|# y: predicted labels (N).|$

\STATE    $\color{agreen}\verb|# cc: cluster centers (num_classes, M, dim).|$
\STATE    $\color{agreen}\verb|# L: clustering results.|$
\STATE    $\color{agreen}\verb|# | \mathtt{\mu} \verb|: momentum coefficient.|$
\STATE
\STATE    $\verb|def | \_\verb|assigning|\_\verb|subclass|\_\verb|labels(x, |\hat{\texttt{y}} \verb|, y):|$
\STATE    $\qquad\color{agreen}\verb|# selected features, subclass labels.|$
\STATE    $\qquad\color{agreen}\verb|# selected cluster center embbedings, cluster center labels.|$
\STATE    $\qquad\texttt{X}_{o} \verb|,| \texttt{y}_{o} \verb|,| \widetilde{\texttt{X}}_{o} \verb|,| \widetilde{\texttt{y}}_{o} = \verb|[],[],[],[] |$
\STATE    $\qquad\color{agreen}\verb|# Record of new cluster centers for this iteration, see Eq.(5).|$
\STATE    $\qquad\verb|ncc = | {\color{apurple}\texttt{zeros}}\verb|(nc, M, dim)|$
\STATE    $\qquad\verb|this|\_\verb|class = | {\color{apurple}\texttt{unique}}\verb|(this_y)|$
\STATE    $\qquad\verb|for idx in this|\_\verb|class:|$
\STATE    $\qquad\qquad\verb|indices = (| \hat{\texttt{y}} \verb| == idx).nonzero()|$
\STATE    $\qquad\qquad\color{agreen}\verb|# select cluster centers with idx.|$
\STATE    $\qquad\qquad\verb|pc = |{\color{ablue}\texttt{select}}\verb|(cc, idx)|$
\STATE    $\qquad\qquad\color{agreen}\verb|# select features with indices.|$
\STATE    $\qquad\qquad\verb|xc = |{\color{ablue}\texttt{select}}\verb|(x, indices)|$
\STATE    $\qquad\qquad\verb|yc = |{\color{ablue}\texttt{select}}\verb|(| \hat{\texttt{y}} \verb|, indices)|$
\STATE    $\qquad\qquad\verb|PS = |{\color{apurple}\texttt{mm}}\verb|(xc, pc.T)|$
\STATE    $\qquad\qquad\verb|PS = |{\color{apurple}\texttt{softmax}}\verb|(PS, 1)|$
\STATE    $\qquad\qquad\color{agreen}\verb|# Sinkhorn-Knopp algorithm.|$
\STATE    $\qquad\qquad\color{apurple}\verb|online|\_\verb|clustering() |$
\STATE    $\qquad\qquad\verb|yc = yc * M|$
\STATE    $\qquad\qquad\verb|yc = yc + L |$
\STATE    $\qquad\qquad\color{agreen}\verb|# Averageing xc tensor according to L.|$
\STATE    $\qquad\qquad\verb|ncc[idx] = |{\color{apurple}\verb|scatter|\_\verb|mean|}\verb|(xc, L, dim=0, dim_size=M)|$
\STATE    $\qquad\qquad\color{agreen}\verb|# append to output variables.|$
\STATE    $\qquad\qquad\texttt{X}_{o} \verb| = | {\color{ablue}\verb|append|}\verb|(| \texttt{X}_{o}\verb|, xc)|$
\STATE    $\qquad\qquad\texttt{y}_{o} \verb| = | {\color{ablue}\verb|append|}\verb|(| \texttt{y}_{o}\verb|, yc)|$
\STATE    $\qquad\qquad\widetilde{\texttt{X}}_{o} \verb| = | {\color{ablue}\verb|append|}\verb|(| \widetilde{\texttt{X}}_{o}\verb|, pc)|$
\STATE    $\qquad\qquad\widetilde{\texttt{y}}_{o} \verb| = | {\color{ablue}\verb|append|}\verb|(| \widetilde{\texttt{y}}_{o}\verb|, idx.| {\color{apurple}\verb|repeat|} \verb|(M) * M| + {\color{apurple}\verb|tensor|}\verb|(|{\color{apurple}\verb|list|}\verb|(|{\color{apurple}\verb|range|}\verb|(M))))|$
\STATE    $\qquad\verb|return |\texttt{X}_{o} \verb|,| \texttt{y}_{o} \verb|,| \widetilde{\texttt{X}}_{o} \verb|,| \widetilde{\texttt{y}}_{o}$
\STATE
\STATE    $\verb|def |\_\verb|update|\_\verb|operation()|$:
\STATE    $\qquad\verb|cc = cc * | \mathtt{\mu} \verb| + ncc * (1 - | \mathtt{\mu}\verb|)|$
\STATE    $\qquad\verb|cc = | {\color{apurple}\verb|normalize|}\verb|((cc, p=2, dim=2))|$
\STATE    $\qquad\verb|return cc|$
\end{algorithmic}
\end{algorithm*}

\begin{algorithm*}[h!]
\caption{Pseudo-code of clustering based point cloud segmentation learning - Part II. }
\label{alg:algorithm2}
\begin{algorithmic}[0] 
\STATE    $\color{agreen}\verb|# temperature: scalar temperature parameter|$
\STATE    $\color{agreen}\verb|# | \texttt{X}_{i} \verb|: selected features|$
\STATE    $\color{agreen}\verb|# | \texttt{y}_{i} \verb|: selected subclass labels|$
\STATE    $\color{agreen}\verb|# | \widetilde{\texttt{X}}_{i} \verb|: cluster center embbedings|$
\STATE    $\color{agreen}\verb|# | \widetilde{\texttt{y}}_{i} \verb|: cluster center labels|$
\STATE
\STATE    $\verb|def | \_\verb|pcc|\_\verb|contrastive(|\texttt{X}_{i} \verb|,| \texttt{y}_{i} \verb|,| \widetilde{\texttt{X}}_{i} \verb|,| \widetilde{\texttt{y}}_{i}\verb|):|$
\STATE    $\qquad\verb|anchor|\_\verb|label = |\texttt{y}_{i}\verb|.view(-1, 1)|$
\STATE    $\qquad\verb|contrast|\_\verb|label = |\widetilde{\texttt{y}}_{i}\verb|.view(-1, 1)|$
\STATE    $\qquad\verb|anchor|\_\verb|feature = | \texttt{X}_{i}$
\STATE    $\qquad\verb|contrast|\_\verb|feature = | \widetilde{\texttt{X}}_{i}$
\STATE
\STATE    $\qquad\verb|mask = |{\color{apurple}\texttt{eq}}\verb|(anchor_label, contrast_label.T)|$

\STATE    $\qquad\verb|anchor_dot_contrast = |{\color{apurple}\texttt{div}}\verb|( |{\color{apurple}\texttt{matmul}}\verb|(anchor_feature, |$
\STATE    $\qquad\qquad\qquad\qquad\qquad\qquad\qquad\quad\verb|contrast_feature.T), temperature) |$
\STATE    $\qquad\verb|logits_max, _ = |{\color{apurple}\texttt{max}}\verb|(anchor_dot_contrast, dim=1, keepdim=True)|$
\STATE    $\qquad\color{agreen}\verb|# To avoid the numerical overflow|$
\STATE    $\qquad\verb|logits = anchor_dot_contrast - logits_max.detach() |$
\STATE
\STATE    $\qquad\color{agreen}\verb|# neg_logits mean the sum of logits of all negative pairs|$
\STATE    $\qquad\verb|neg_mask = 1 - mask|$
\STATE    $\qquad\verb|neg_logits = |{\color{apurple}\texttt{exp}}\verb|(logits) * neg_mask|$
\STATE    $\qquad\verb|neg_logits = neg_logits.sum(1, keepdim=True) |$
\STATE
\STATE    $\qquad\color{agreen}\verb|# exp_logits mean the logit of each sample pair|$
\STATE    $\qquad\verb|exp_logits = |{\color{apurple}\texttt{exp}}\verb|(logits) |$
\STATE    $\qquad\verb|log_prob = logits - |{\color{apurple}\texttt{log}}\verb|(exp_logits + neg_logits)|$
\STATE    $\qquad\verb|mean_log_prob_pos = (mask * log_prob).sum(1) / mask.sum(1)|$
\STATE
\STATE    $\qquad\verb|loss = - (temperature / base_temperature) * mean_log_prob_pos|$
\STATE    $\qquad\verb|loss = loss.mean()|$
\STATE    $\qquad\verb|return loss|$
\end{algorithmic}
\end{algorithm*}

\newpage

\newpage

\begin{table*}[h!]
\vspace{-5pt}
\begin{center}
    \captionsetup{width=.98\textwidth}
\caption{$_{\!}$\textbf{Quantitative$_{\!}$ results$_{\!}$} on$_{\!}$ SemanticKITTI$_{\!}$~\cite{behley2019semantickitti}$_{\!}$ single-scan$_{\!}$ challenge$_{\!}$ \texttt{test}$_{\!}$ (\S\ref{sec:ex1}) - Part I. mIoU (\%) and IoUs (\%) are reported.}
\vspace{-5pt}
\label{table:StaticUrbantestappendix}
\setlength\tabcolsep{12pt}
\renewcommand\arraystretch{1.18}
\resizebox{\linewidth}{!}{
\begin{tabular}{|r||c|cccccccccc|}
\thickhline
\rowcolor{mygray}
Method&mIoU&\rotatebox{90}{road}&\rotatebox{90}{sidewalk}&\rotatebox{90}{parking}&\rotatebox{90}{other-ground}&\rotatebox{90}{building}&\rotatebox{90}{car}&\rotatebox{90}{truck}&\rotatebox{90}{bicycle}&\rotatebox{90}{motorcycle}&\rotatebox{90}{other-vehicle}\\
\hline
\hline
TangentConv\pub{CVPR18}\cite{tatarchenko2018tangent}&40.9&83.9&63.9&33.4&15.4&83.4&90.8&15.2&2.7&16.5&12.1\\
SqueezeSegV2\pub{ICRA19}\cite{wu2019squeezesegv2}&39.7&88.6&67.6&45.8&17.7&73.7&81.8&13.4&18.5&17.9&14.0\\
DarkNet53\pub{ICCV19}\cite{behley2019semantickitti}&49.9&91.8&74.6&64.8&27.9&84.1&86.4&25.5&24.5&32.7&22.6\\
Rangenet++\pub{IROS19}\cite{milioto2019rangenet++}&52.2&91.8&75.2&65.0&27.8&87.4&91.4&25.7&25.7&34.4&23.0\\
3D-MiniNet\pub{IROS20}\cite{alonso2020MiniNet3D}&55.8&91.6&74.5&64.2&25.4&89.4&90.5&28.5&42.3&42.1&29.4\\
PointASNL\pub{CVPR20}\cite{Yan_2020_CVPR}&46.8&87.4&74.3&24.3&1.8&83.1&87.9&39.0&0.0&25.1&29.2\\
PolarNet\pub{CVPR20}\cite{zhang2020polarnet}&54.3&90.8&74.4&61.7&21.7&90.0&93.8&22.9&40.3&30.1&28.5\\
RandLA-Net\pub{CVPR20}\cite{hu2020randla}&55.9&90.5&74.0&61.8&24.5&89.7&94.2&43.9&29.8&32.2&39.1\\
SqueezeSegV3\pub{ECCV20}\cite{xu2020squeezesegv3}&55.9&91.7&74.8&63.4&26.4&89.0&92.5&29.6&38.7&36.5&33.0\\
SalsaNext\pub{ISVC20}\cite{cortinhal2020salsanext}&59.5&91.7&75.8&63.7&29.1&90.2&91.9&38.9&48.3&38.6&31.9\\
FusionNet\pub{ECCV20}\cite{zhang2020deep}&61.3&91.8&77.1&68.8&30.8&92.5&95.3&41.8&47.5&37.7&34.5\\
JS3C-Net\pub{AAAI21}\cite{yan2021sparse}&66.0&88.9&72.1&61.9&31.9&92.5&95.8&\bf{54.3}&59.3&52.9&46.0\\
AF2S3Net\pub{CVPR21}\cite{cheng20212}&69.7&91.3&72.5&68.8&\bf{53.5}&87.9&94.5&39.2&65.4&\bf{86.8}&41.1\\
RPVNet\pub{ICCV21}\cite{xu2021rpvnet}&70.3&\bf{93.4}&\bf{80.7}&70.3&33.3&\bf{93.5}&\bf{97.6}&44.2&\bf{68.4}&68.7&\bf{61.1}\\
PVKD\pub{CVPR22}\cite{hou2022point}&\bf{71.4}&91.8&77.5&\bf{70.9}&41.0&92.4&97.0&53.5&67.9&69.3&60.2\\
\hline
KPConv\pub{ICCV19}\cite{thomas2019kpconv}&58.8&88.8&72.7&61.3&31.6&90.5&96.0&33.4&30.2&42.5&44.3\\
KPConv $+~\textbf{\texttt{Ours}}$ &61.0&89.9&75.0&63.4&34.3&91.4&88.8&49.0& 45.0&46.6&45.5\\
      \cdashline{1-12}[1pt/1pt]
SPVNAS$_{\text{10.8M}}$\pub{ECCV20}\cite{tang2020searching}&62.3 & 89.6&73.8&63.2&29.1&90.9&96.7&50.9&40.6&42.1&51.3\\
SPVNAS$_{\text{10.8M}}$ +~\textbf{\texttt{Ours}} &64.3&89.6&73.9&64.0&28.8&91.4&96.7&48.0&48.9&50.5&51.0\\
      \cdashline{1-12}[1pt/1pt]
Cylinder3D\pub{CVPR21}\cite{zhu2021cylindrical} &67.8&91.4&75.5&65.1&32.3&91.0&97.1&50.8&67.6&64.0&58.6\\
Cylinder3D +~\texttt{\textbf{Ours}} &70.4&91.7&77.2&66.1&34.1&92.3&97.0&51.9&\bf{68.4}&65.8&58.8\\
\hline
\end{tabular}}
\vspace{-20pt}
\end{center}
\end{table*}

\begin{table*}[h!]
\begin{center}
    \captionsetup{width=.98\textwidth}
\caption{$_{\!}$\textbf{Quantitative$_{\!}$ results$_{\!}$} on$_{\!}$ SemanticKITTI$_{\!}$~\cite{behley2019semantickitti}$_{\!}$ single-scan$_{\!}$ challenge$_{\!}$ \texttt{test}$_{\!}$ (\S\ref{sec:ex1}) - Part II. mIoU (\%) and IoUs (\%) are reported.
}
\vspace{-5pt}
\label{table:StaticUrbantestappendix2}
\setlength\tabcolsep{13pt}
\renewcommand\arraystretch{1.18}
\resizebox{\linewidth}{!}{
\begin{tabular}{|r||c|ccccccccc|}
\thickhline
\rowcolor{mygray}
Method&mIoU&\rotatebox{90}{vegetation}&\rotatebox{90}{trunk}&\rotatebox{90}{terrain}&\rotatebox{90}{person}&\rotatebox{90}{bicyclist}&\rotatebox{90}{motorcyclist}&\rotatebox{90}{fence}&\rotatebox{90}{pole}&\rotatebox{90}{traffic-sign}\\
\hline
\hline
TangentConv\pub{CVPR18}\cite{tatarchenko2018tangent}&40.9&79.5&49.3&58.1&23.0&28.4&8.1&49.0&35.8&28.5\\
SqueezeSegV2\pub{ICRA19}\cite{wu2019squeezesegv2}&39.7&71.8&35.8&60.2&20.1&25.1&3.9&41.1&20.2&26.3\\
DarkNet53\pub{ICCV19}\cite{behley2019semantickitti}&49.9&78.3&50.1&64.0&36.2&33.6&4.7&55.0&38.9&52.2\\
Rangenet++\pub{IROS19}\cite{milioto2019rangenet++}&52.2&80.5&55.1&64.6&38.3&38.8&4.8&58.6&47.9&55.9\\
3D-MiniNet\pub{IROS20}\cite{alonso2020MiniNet3D}&55.8&82.8&60.8&66.7&47.8&44.1&14.5&60.8&48.0&56.6\\
PointASNL\pub{CVPR20}\cite{Yan_2020_CVPR}&46.8&84.1&52.2&70.6&34.2&57.6&0.0&43.9&57.8&36.9\\
PolarNet\pub{CVPR20}\cite{zhang2020polarnet}&54.3&84.0&65.5&67.8&43.2&40.2&5.6&61.3&51.8&57.5\\
RandLA-Net\pub{CVPR20}\cite{hu2020randla}&55.9&83.8&63.6&68.6&48.4&47.4&9.4&60.4&51.0&50.7\\
SqueezeSegV3\pub{ECCV20}\cite{xu2020squeezesegv3}&55.9&82.0&58.7&65.4&45.6&46.2&20.1&59.4&49.6&58.9\\
SalsaNext\pub{ISVC20}\cite{cortinhal2020salsanext}&59.5&81.8&63.6&66.5&60.2&59.0&19.4&64.2&54.3&62.1\\
FusionNet\pub{ECCV20}\cite{zhang2020deep}&61.3&84.5&69.8&68.5&59.5&56.8&11.9&69.4&60.4&66.5\\
JS3C-Net\pub{AAAI21}\cite{yan2021sparse}&66.0&84.5&69.8&67.9&69.5&65.4&39.9&70.8&60.7&68.7\\
AF2S3Net\pub{CVPR21}\cite{cheng20212}&69.7&70.2&68.5&53.7&\bf{80.7}&\bf{80.4}&\bf{74.3}&63.2&61.5&71.0\\
RPVNet\pub{ICCV21}\cite{xu2021rpvnet}&70.3&86.5&\bf{75.1}&71.7&75.9&74.4&43.4&\bf{72.1}&64.8&61.4\\
PVKD\pub{CVPR22}\cite{hou2022point}&\bf{71.4}&86.5&73.8&\bf{71.9}&75.1&73.5&50.5&69.4&64.9&61.4\\
\hline
KPConv\pub{ICCV19}\cite{thomas2019kpconv}&58.8&84.8&69.2&69.1&61.5&61.6&11.8&64.2&56.4&47.4\\
KPConv $+~\textbf{\texttt{Ours}}$ &61.0&72.0&56.5&68.8&59.4&60.1& 36.4&66.1&49.5&60.4\\
      \cdashline{1-11}[1pt/1pt]
SPVNAS$_{\text{10.8M}}$\pub{ECCV20}\cite{tang2020searching}&62.3 &85.5&70.3&69.8&60.4&62.8&21.8&65.3&57.6&62.0\\
SPVNAS$_{\text{10.8M}}$ +~\textbf{\texttt{Ours}} &64.3&85.3&72.1&69.1&67.1&70.5&23.2&67.0&60.7&64.5\\
      \cdashline{1-11}[1pt/1pt]
Cylinder3D\pub{CVPR21}\cite{zhu2021cylindrical} &67.8&85.4&71.8&68.5&73.9&67.9&36.0&66.5&62.6&65.6\\
Cylinder3D +~\texttt{\textbf{Ours}} &70.4&\bf{86.7}&73.5&71.7&69.6&70.1&54.6&70.8&\bf{65.1}&\bf{71.6}\\
\hline
\end{tabular}}
\vspace{-25pt}
\end{center}
\end{table*}

\newpage

\begin{table*}[h!]
\begin{center}
    \captionsetup{width=.98\textwidth}
\caption{$_{\!}$\textbf{Quantitative$_{\!}$ results$_{\!}$} on$_{\!}$ S3DIS$_{\!}$~\cite{armeni20163d}$_{\!}$ Area-5$_{\!}$ (\S\ref{sec:ex2}) - Part I. mIoU (\%) and IoUs (\%) are reported.}
\label{table:StaticIndoorvalappendix}
\setlength\tabcolsep{14pt}
\renewcommand\arraystretch{1.18}
\resizebox{\linewidth}{!}{
\begin{tabular}{|r||ccc|ccccccc|}
\thickhline
\rowcolor{mygray}
Method&mIoU&mAcc&OA&\rotatebox{90}{ceiling}&\rotatebox{90}{floor}&\rotatebox{90}{wall}&\rotatebox{90}{beam}&\rotatebox{90}{column}&\rotatebox{90}{window}&\rotatebox{90}{door}\\
\hline
\hline
PointNet\pub{CVPR17}\cite{Qi_2017_CVPR}& 41.1&49.0&-& 88.8&97.3&69.8&0.1&3.9&46.3&10.8\\
SegCloud\pub{3DV17}\cite{tchapmi2017segcloud}&48.9&57.4&-&90.1&96.1&69.9&0.0&18.4&38.4&23.1\\
TangentConv\pub{CVPR18}\cite{tatarchenko2018tangent}&52.6&62.2&-&90.5&97.7&74.0&0.0&20.7&39.0&31.3\\
PointCNN\pub{NeurIPS18}\cite{li2018pointcnn}&57.3&63.9&85.9&92.3&98.2&79.4&0.0&17.6&22.8&62.1\\
SPGraph\pub{CVPR18}\cite{landrieu2018large}&58.0&66.5&86.4&89.4&96.9&78.1&0.0&42.8&48.9&61.6\\
PCCN\pub{CVPR18}\cite{wang2018deep}&58.3&-&67.0&92.3&96.2&75.9&0.3&6.0&69.5&63.5\\
HPEIN\pub{ICCV19}\cite{jiang2019hierarchical}&61.9&68.3&87.2&91.5&98.2&81.4&0.0&23.3&65.3&40.0\\
PAT\pub{CVPR19}\cite{yang2019modeling}&60.1&70.8&-&93.0&98.5&72.3&1.0&41.5&85.1&38.2\\
PointWeb\pub{CVPR19}\cite{zhao2019pointweb}&60.3&66.6&87.0&92.0&98.5&79.4&0.0&21.1&59.7&34.8\\
MinkowskiNet\pub{CVPR19}\cite{choy20194d}&65.4&71.7&-&91.8&98.7&86.2&0.0&34.1&48.9&62.4\\
SCF-Net\pub{CVPR21}\cite{fan2021scf}&63.8&-&-&-&-&-&-&-&-&-\\
BAAF-Net\pub{CVPR21}\cite{qiu2021semantic}&65.4&73.1&88.9&-&-&-&-&-&-&-\\
CGA-Net\pub{CVPR21}\cite{lu2021cga}&68.6&-&-&94.5&98.3&83.0&0.0&25.3&59.6&71.0\\
Stratified Trans.\pub{CVPR22}\cite{lai2022stratified}&72.0&78.1&91.5&-&-&-&-&-&-&-\\
PTV2\pub{NeurIPS22}\cite{wupoint}&\bf{72.6}&78.0&\bf{91.6}&-&-&-&-&-&-&-\\
\hline
KPConv\pub{ICCV19}\cite{thomas2019kpconv}&67.1&72.8&-&92.8&97.3&82.4&0.0&23.9&58.0&69.0\\
KPConv$+~\textbf{\texttt{Ours}}$&69.0&76.2&90.5&95.7&98.3&84.0&0.0&30.7&66.7&77.6\\
      \cdashline{1-11}[1pt/1pt]
PTV1\pub{ICCV21}\cite{zhao2021point}&70.4&76.5&90.8&94.0&98.5&86.3&0.0&38.0&63.4&74.3\\
PTV1$+$CBL\pub{CVPR22}\cite{tang2022contrastive}&71.6&77.9&91.2&-&-&-&-&-&-&-\\
PTV1$+~\textbf{\texttt{Ours}}$&72.2&\bf{79.6}&91.2&94.2&98.4&88.1&0.0&49.3&65.3&79.4\\
\hline
\end{tabular}}
\end{center}
\end{table*}

\begin{table*}[h!]
\begin{center}
    \captionsetup{width=.98\textwidth}
\caption{$_{\!}$\textbf{Quantitative$_{\!}$ results$_{\!}$} on$_{\!}$ S3DIS$_{\!}$~\cite{armeni20163d}$_{\!}$ Area-5$_{\!}$ (\S\ref{sec:ex2}) - Part II. mIoU (\%) and IoUs (\%) are reported.}
\label{table:StaticIndoorvalappendix2}
\setlength\tabcolsep{15pt}
\renewcommand\arraystretch{1.18}
\resizebox{\linewidth}{!}{
\begin{tabular}{|r||ccc|cccccc|}
\thickhline
\rowcolor{mygray}
Method&mIoU&mAcc&OA&\rotatebox{90}{table}&\rotatebox{90}{chair}&\rotatebox{90}{sofa}&\rotatebox{90}{bookcase}&\rotatebox{90}{board}&\rotatebox{90}{clutter}\\
\hline
\hline
PointNet\pub{CVPR17}\cite{Qi_2017_CVPR}& 41.1&49.0&-& 52.6&58.9&40.3&5.9&26.4&33.3\\
SegCloud\pub{3DV17}\cite{tchapmi2017segcloud}&48.9&57.4&-&70.4&75.9&40.9&58.4&13.0&41.6\\
TangentConv\pub{CVPR18}\cite{tatarchenko2018tangent}&52.6&62.2&-&77.5&69.4&57.3&38.5&48.8&39.8\\
PointCNN\pub{NeurIPS18}\cite{li2018pointcnn}&57.3&63.9&85.9&74.4&80.6&31.7&66.7&62.1&56.7\\
SPGraph\pub{CVPR18}\cite{landrieu2018large}&58.0&66.5&86.4&84.7&75.4&69.8&52.6&2.1&52.2\\
PCCN\pub{CVPR18}\cite{wang2018deep}&58.3&-&67.0&66.9&65.6&47.3&68.9&59.1&46.2\\
HPEIN\pub{ICCV19}\cite{jiang2019hierarchical}&61.9&68.3&87.2&75.5&87.7&58.5&67.8&65.6&49.4\\
PAT\pub{CVPR19}\cite{yang2019modeling}&60.1&70.8&-&57.7&83.6&48.1&67.0&61.3&33.6\\
PointWeb\pub{CVPR19}\cite{zhao2019pointweb}&60.3&66.6&87.0&76.3&88.3&46.9&69.3&64.9&52.5\\
MinkowskiNet\pub{CVPR19}\cite{choy20194d}&65.4&71.7&-&81.6&89.8&47.2&74.9&74.4&58.6\\
SCF-Net\pub{CVPR21}\cite{fan2021scf}&63.8&-&-&-&-&-&-&-&-\\
BAAF-Net\pub{CVPR21}\cite{qiu2021semantic}&65.4&73.1&88.9&-&-&-&-&-&-\\
CGA-Net\pub{CVPR21}\cite{lu2021cga}&68.6&-&-&82.6&92.2&77.7&76.4&69.5&61.5\\
Stratified Trans.\pub{CVPR22}\cite{lai2022stratified}&72.0&78.1&91.5&-&-&-&-&-&-\\
PTV2\pub{NeurIPS22}\cite{wupoint}&\bf{72.6}&78.0&\bf{91.6}&-&-&-&-&-&-\\
\hline
KPConv\pub{ICCV19}\cite{thomas2019kpconv}&67.1&72.8&-&81.5&91.0&75.4&75.3&66.7&58.9\\
KPConv$+~\textbf{\texttt{Ours}}$&69.0&76.2&90.8&79.9&91.0&70.3&76.7&63.0&63.6\\
      \cdashline{1-10}[1pt/1pt]
PTV1\pub{ICCV21}\cite{zhao2021point}&70.4&76.5&90.8&89.1&82.4&74.3&80.2&76.0&59.3\\
PTV1$+$CBL\pub{CVPR22}\cite{tang2022contrastive}&71.6&77.9&91.2&-&-&-&-&-&-\\
PTV1$+~\textbf{\texttt{Ours}}$&72.2&\bf{79.6}&91.2&89.4&82.2&74.8&77.6&81.0&58.7\\
\hline
\end{tabular}}
\end{center}
\end{table*}

\newpage

\begin{table*}[h!]
\begin{center}
    \captionsetup{width=.98\textwidth}
\caption{$_{\!}$\textbf{Quantitative$_{\!}$ results$_{\!}$} on$_{\!}$ SemanticKITTI$_{\!}$~\cite{behley2019semantickitti}$_{\!}$ multi-scan$_{\!}$ challenge$_{\!}$ \texttt{test}$_{\!}$  (\S\ref{sec:ex3}) - Part I. mIoU (\%) and IoUs (\%) are reported.
}
\label{table:4DSegmentationtest1appendix}
\setlength\tabcolsep{7.5pt}
\renewcommand\arraystretch{1.18}
\resizebox{\linewidth}{!}{
\begin{tabular}{|r||c|ccccccccccccc|}
\thickhline
\rowcolor{mygray}
Method&mIoU&\rotatebox{90}{road}&\rotatebox{90}{sidewalk}&\rotatebox{90}{parking}&\rotatebox{90}{other-ground}&\rotatebox{90}{building}&\rotatebox{90}{car}&\rotatebox{90}{moving car}&\rotatebox{90}{truck}&\rotatebox{90}{moving truck}&\rotatebox{90}{bicycle}&\rotatebox{90}{motorcycle}&\rotatebox{90}{other-vehicle}&\rotatebox{90}{moving other-vehicle}\\
\hline
\hline
TangentConv\pub{CVPR18}\cite{tatarchenko2018tangent}&34.1&83.9&64.0&38.3&15.3&85.8&84.9&40.3&21.1&1.1&2.0&18.2&18.5&6.4\\
DarkNet53\pub{ICCV19}\cite{behley2019semantickitti}&41.6&91.6&75.3&64.9&27.5&85.2&84.1&61.5&20.0&14.1&30.4&32.9&20.7&15.2\\
TemporalLidarSeg\pub{3DV20}\cite{duerr2020lidar}&47.0&\bf{91.8}&75.8&59.6&23.2&89.8&92.1&68.2&39.2&2.1&47.7&40.9&35.0&12.4\\
SpSeqnet\pub{CVPR20}\cite{shi2020spsequencenet}&43.1&90.1&73.9&57.6&27.1&91.2&88.5&53.2&29.2&\bf{41.2}&24.0&26.2&22.7&\bf{26.2}\\
\hline
KPConv\pub{ICCV19}\cite{thomas2019kpconv}&51.2&86.5&70.5&58.4&26.7&90.8&93.7&69.4&42.5&5.8&44.9&47.2&38.6&4.7\\
KPConv$+~\textbf{\texttt{Ours}}$&53.2&90.4&75.2&62.1&25.1&91.8&\bf{95.8}&75.2&\bf{43.8}&4.1&67.2&63.1&\bf{44.2}&0.7\\
      \cdashline{1-15}[1pt/1pt]
Cylinder3D\pub{CVPR21}\cite{zhu2021cylindrical}&52.5&90.7&74.5&65.0&\bf{32.3}&\bf{92.6}&94.6&74.9&41.3&0.0&\bf{67.6}&\bf{63.8}&38.8&0.1\\
Cylinder3D$+~\textbf{\texttt{Ours}}$&\bf{54.7}&91.4&\bf{76.9}&\bf{66.1}&27.8&91.4&95.3&\bf{81.7}&42.7&11.9&55.9&52.9&38.7&11.2\\
\hline
\end{tabular}}
\end{center}
\end{table*}


\begin{table*}[h!]
\begin{center}
    \captionsetup{width=.98\textwidth}
\caption{$_{\!}$\textbf{Quantitative$_{\!}$ results$_{\!}$} on$_{\!}$ SemanticKITTI$_{\!}$~\cite{behley2019semantickitti}$_{\!}$ multi-scan$_{\!}$ challenge$_{\!}$ \texttt{test}$_{\!}$  (\S\ref{sec:ex3}) - Part II. mIoU (\%) and IoUs (\%) are reported.
}
\label{table:4DSegmentationtest1appendix2}
\setlength\tabcolsep{8.5pt}
\renewcommand\arraystretch{1.18}
\resizebox{\linewidth}{!}{
\begin{tabular}{|r||c|cccccccccccc|}
\thickhline
\rowcolor{mygray}
Method&mIoU&\rotatebox{90}{vegetation}&\rotatebox{90}{trunk}&\rotatebox{90}{terrain}&\rotatebox{90}{person}&\rotatebox{90}{moving person}&\rotatebox{90}{bicyclist}&\rotatebox{90}{moving bicyclist}&\rotatebox{90}{motorcyclist}&\rotatebox{90}{moving motorcyclist}&\rotatebox{90}{fence}&\rotatebox{90}{pole}&\rotatebox{90}{traffic-sign}\\
\hline
\hline
TangentConv\pub{CVPR18}\cite{tatarchenko2018tangent}&34.1&79.5& 43.2& 56.7&1.6&1.9&0.0&30.1&0.0&42.2&49.1&36.4&31.2\\
DarkNet53\pub{ICCV19}\cite{behley2019semantickitti}&41.6&78.4& 50.7& 64.8&7.5&0.2&0.0&28.9&0.0&37.8&56.5&38.1&53.3\\
TemporalLidarSeg\pub{3DV20}\cite{duerr2020lidar}&47.0&82.3&62.5&64.7&14.4&40.4&0.0&42.8&0.0&12.9&63.8&52.6&60.4\\
SpSeqnet\pub{CVPR20}\cite{shi2020spsequencenet}&43.1&84.0& 66.0&65.7&6.3&36.2&0.0&2.3&0.0&0.1&66.8&50.8&48.7 \\
\hline
KPConv\pub{ICCV19}\cite{thomas2019kpconv}&51.2&84.6&70.3&66.0&\bf{21.6}&\bf{67.5}&0.0&67.4&0.0&\bf{47.2}&64.5&57.0&53.9\\
KPConv$+~\textbf{\texttt{Ours}}$&53.2&85.4&71.1&69.3&10.7&72.1&0.0&\bf{68.5}&\bf{9.9}&9.9&\bf{67.5}&62.6&64.6\\
      \cdashline{1-14}[1pt/1pt]
Cylinder3D\pub{CVPR21}\cite{zhu2021cylindrical}&52.5&85.8&72.0&68.9&12.5&65.7&\bf{1.7}&68.3&0.2&11.9&66.0&63.1&61.4\\
Cylinder3D$+~\textbf{\texttt{Ours}}$&\bf{54.7}&\bf{86.5}&\bf{72.7}&\bf{71.6}&15.5&61.8&0.0&68.2&3.0&46.0&66.1&\bf{64.0}&\bf{68.0}\\
\hline
\end{tabular}}
\end{center}
\end{table*}

\newpage

\newpage

\newpage

\clearpage
\clearpage

{\small
\bibliographystyle{ieee_fullname}
\bibliography{egbib}

\begin{thebibliography}{100}\itemsep=-1pt

\bibitem{tchapmi2017segcloud}
Lyne Tchapmi, Christopher Choy, Iro Armeni, JunYoung Gwak, and Silvio Savarese.
\newblock Segcloud: Semantic segmentation of 3d point clouds.
\newblock In {\em 3DV}, 2017.

\bibitem{wu2018squeezeseg}
Bichen Wu, Alvin Wan, Xiangyu Yue, and Kurt Keutzer.
\newblock Squeezeseg: Convolutional neural nets with recurrent crf for
  real-time road-object segmentation from 3d lidar point cloud.
\newblock In {\em ICRA}, 2018.

\bibitem{graham20183d}
Benjamin Graham, Martin Engelcke, and Laurens Van Der~Maaten.
\newblock {3D} semantic segmentation with submanifold sparse convolutional
  networks.
\newblock In {\em CVPR}, 2018.

\bibitem{meng2019vv}
Hsien-Yu Meng, Lin Gao, Yu-Kun Lai, and Dinesh Manocha.
\newblock Vv-net: Voxel vae net with group convolutions for point cloud
  segmentation.
\newblock In {\em ICCV}, 2019.

\bibitem{milioto2019rangenet++}
Andres Milioto, Ignacio Vizzo, Jens Behley, and Cyrill Stachniss.
\newblock Rangenet++: Fast and accurate lidar semantic segmentation.
\newblock In {\em IROS}, 2019.

\bibitem{choy20194d}
Christopher Choy, JunYoung Gwak, and Silvio Savarese.
\newblock 4d spatio-temporal convnets: Minkowski convolutional neural networks.
\newblock In {\em CVPR}, 2019.

\bibitem{wu2019squeezesegv2}
Bichen Wu, Xuanyu Zhou, Sicheng Zhao, Xiangyu Yue, and Kurt Keutzer.
\newblock Squeezesegv2: Improved model structure and unsupervised domain
  adaptation for road-object segmentation from a lidar point cloud.
\newblock In {\em ICRA}, 2019.

\bibitem{xu2020squeezesegv3}
Chenfeng Xu, Bichen Wu, Zining Wang, Wei Zhan, Peter Vajda, Kurt Keutzer, and
  Masayoshi Tomizuka.
\newblock Squeezesegv3: Spatially-adaptive convolution for efficient
  point-cloud segmentation.
\newblock In {\em ECCV}, 2020.

\bibitem{zhang2020polarnet}
Yang Zhang, Zixiang Zhou, Philip David, Xiangyu Yue, Zerong Xi, Boqing Gong,
  and Hassan Foroosh.
\newblock Polarnet: An improved grid representation for online lidar point
  clouds semantic segmentation.
\newblock In {\em CVPR}, 2020.

\bibitem{cortinhal2020salsanext}
Tiago Cortinhal, George Tzelepis, and Eren~Erdal Aksoy.
\newblock Salsanext: fast, uncertainty-aware semantic segmentation of lidar
  point clouds for autonomous driving.
\newblock {\em arXiv preprint arXiv:2003.03653}, 2020.

\bibitem{hu2021vmnet}
Zeyu Hu, Xuyang Bai, Jiaxiang Shang, Runze Zhang, Jiayu Dong, Xin Wang,
  Guangyuan Sun, Hongbo Fu, and Chiew-Lan Tai.
\newblock Vmnet: Voxel-mesh network for geodesic-aware 3d semantic
  segmentation.
\newblock In {\em ICCV}, 2021.

\bibitem{engelmann2020dilated}
Francis Engelmann, Theodora Kontogianni, and Bastian Leibe.
\newblock Dilated point convolutions: On the receptive field size of point
  convolutions on 3d point clouds.
\newblock In {\em ICRA}, 2020.

\bibitem{hua2018pointwise}
Binh-Son Hua, Minh-Khoi Tran, and Sai-Kit Yeung.
\newblock Pointwise convolutional neural networks.
\newblock In {\em CVPR}, 2018.

\bibitem{zhao2019pointweb}
Hengshuang Zhao, Li Jiang, Chi-Wing Fu, and Jiaya Jia.
\newblock Pointweb: Enhancing local neighborhood features for point cloud
  processing.
\newblock In {\em CVPR}, 2019.

\bibitem{zhang2019shellnet}
Zhiyuan Zhang, Binh-Son Hua, and Sai-Kit Yeung.
\newblock Shellnet: Efficient point cloud convolutional neural networks using
  concentric shells statistics.
\newblock In {\em ICCV}, 2019.

\bibitem{zhu2021cylindrical}
Xinge Zhu, Hui Zhou, Tai Wang, Fangzhou Hong, Yuexin Ma, Wei Li, Hongsheng Li,
  and Dahua Lin.
\newblock Cylindrical and asymmetrical 3d convolution networks for lidar
  segmentation.
\newblock In {\em CVPR}, 2021.

\bibitem{Qi_2017_CVPR}
Charles~R. Qi, Hao Su, Kaichun Mo, and Leonidas~J. Guibas.
\newblock Pointnet: Deep learning on point sets for {3D} classification and
  segmentation.
\newblock In {\em CVPR}, 2017.

\bibitem{qi2017}
Charles~Ruizhongtai Qi, Li Yi, Hao Su, and Leonidas~J. Guibas.
\newblock Pointnet++: Deep hierarchical feature learning on point sets in a
  metric space.
\newblock In {\em NeurIPS}, 2017.

\bibitem{Wu_2019_CVPR}
Wenxuan Wu, Zhongang Qi, and Li Fuxin.
\newblock Pointconv: Deep convolutional networks on 3d point clouds.
\newblock In {\em CVPR}, 2019.

\bibitem{yang2019modeling}
Jiancheng Yang, Qiang Zhang, Bingbing Ni, Linguo Li, Jinxian Liu, Mengdie Zhou,
  and Qi Tian.
\newblock Modeling point clouds with self-attention and gumbel subset sampling.
\newblock In {\em CVPR}, 2019.

\bibitem{hu2020randla}
Qingyong Hu, Bo Yang, Linhai Xie, Stefano Rosa, Yulan Guo, Zhihua Wang, Niki
  Trigoni, and Andrew Markham.
\newblock Randla-net: Efficient semantic segmentation of large-scale point
  clouds.
\newblock In {\em CVPR}, 2020.

\bibitem{landrieu2018large}
Loic Landrieu and Martin Simonovsky.
\newblock Large-scale point cloud semantic segmentation with superpoint graphs.
\newblock In {\em CVPR}, 2018.

\bibitem{wang2018deep}
Shenlong Wang, Simon Suo, Wei-Chiu Ma, Andrei Pokrovsky, and Raquel Urtasun.
\newblock Deep parametric continuous convolutional neural networks.
\newblock In {\em CVPR}, 2018.

\bibitem{ummenhofer2019lagrangian}
Benjamin Ummenhofer, Lukas Prantl, Nils Thuerey, and Vladlen Koltun.
\newblock Lagrangian fluid simulation with continuous convolutions.
\newblock In {\em ICLR}, 2019.

\bibitem{thomas2019kpconv}
Hugues Thomas, Charles~R Qi, Jean-Emmanuel Deschaud, Beatriz Marcotegui,
  Fran{\c{c}}ois Goulette, and Leonidas~J Guibas.
\newblock Kpconv: Flexible and deformable convolution for point clouds.
\newblock In {\em ICCV}, 2019.

\bibitem{zhao2021point}
Hengshuang Zhao, Li Jiang, Jiaya Jia, Philip~HS Torr, and Vladlen Koltun.
\newblock Point transformer.
\newblock In {\em ICCV}, 2021.

\bibitem{mazur2021cloud}
Kirill Mazur and Victor Lempitsky.
\newblock Cloud transformers: A universal approach to point cloud processing
  tasks.
\newblock In {\em ICCV}, 2021.

\bibitem{fan2021point}
Hehe Fan, Yi Yang, and Mohan Kankanhalli.
\newblock Point 4d transformer networks for spatio-temporal modeling in point
  cloud videos.
\newblock In {\em CVPR}, 2021.

\bibitem{knight2008sinkhorn}
Philip~A Knight.
\newblock The sinkhorn--knopp algorithm: convergence and applications.
\newblock {\em SIAM Journal on Matrix Analysis and Applications},
  30(1):261--275, 2008.

\bibitem{cuturi2013sinkhorn}
Marco Cuturi.
\newblock Sinkhorn distances: Lightspeed computation of optimal transport.
\newblock In {\em NeurIPS}, 2013.

\bibitem{asano2019self}
YM Asano, C Rupprecht, and A Vedaldi.
\newblock Self-labelling via simultaneous clustering and representation
  learning.
\newblock In {\em ICLR}, 2019.

\bibitem{tang2020searching}
Haotian Tang, Zhijian Liu, Shengyu Zhao, Yujun Lin, Ji Lin, Hanrui Wang, and
  Song Han.
\newblock Searching efficient 3d architectures with sparse point-voxel
  convolution.
\newblock In {\em ECCV}, 2020.

\bibitem{behley2019semantickitti}
Jens Behley, Martin Garbade, Andres Milioto, Jan Quenzel, Sven Behnke, Cyrill
  Stachniss, and Jurgen Gall.
\newblock Semantickitti: A dataset for semantic scene understanding of lidar
  sequences.
\newblock In {\em ICCV}, 2019.

\bibitem{armeni20163d}
Iro Armeni, Ozan Sener, Amir~R Zamir, Helen Jiang, Ioannis Brilakis, Martin
  Fischer, and Silvio Savarese.
\newblock 3d semantic parsing of large-scale indoor spaces.
\newblock In {\em CVPR}, 2016.

\bibitem{yan2018second}
Yan Yan, Yuxing Mao, and Bo Li.
\newblock Second: Sparsely embedded convolutional detection.
\newblock {\em Sensors}, 18(10):3337, 2018.

\bibitem{lang2019pointpillars}
Alex~H Lang, Sourabh Vora, Holger Caesar, Lubing Zhou, Jiong Yang, and Oscar
  Beijbom.
\newblock Pointpillars: Fast encoders for object detection from point clouds.
\newblock In {\em CVPR}, 2019.

\bibitem{geiger2013vision}
Andreas Geiger, Philip Lenz, Christoph Stiller, and Raquel Urtasun.
\newblock Vision meets robotics: The kitti dataset.
\newblock {\em The International Journal of Robotics Research},
  32(11):1231--1237, 2013.

\bibitem{tatarchenko2018tangent}
Maxim Tatarchenko, Jaesik Park, Vladlen Koltun, and Qian-Yi Zhou.
\newblock Tangent convolutions for dense prediction in 3d.
\newblock In {\em CVPR}, 2018.

\bibitem{lyu2020learning}
Yecheng Lyu, Xinming Huang, and Ziming Zhang.
\newblock Learning to segment 3d point clouds in 2d image space.
\newblock In {\em CVPR}, 2020.

\bibitem{yang2020pfcnn}
Yuqi Yang, Shilin Liu, Hao Pan, Yang Liu, and Xin Tong.
\newblock Pfcnn: convolutional neural networks on 3d surfaces using parallel
  frames.
\newblock In {\em CVPR}, 2020.

\bibitem{Maturana2015}
Daniel Maturana and Sebastian Scherer.
\newblock Voxnet: A {3D} convolutional neural network for real-time object
  recognition.
\newblock In {\em IROS}, 2015.

\bibitem{riegler2017octnet}
Gernot Riegler, Ali Osman~Ulusoy, and Andreas Geiger.
\newblock Octnet: Learning deep 3d representations at high resolutions.
\newblock In {\em CVPR}, 2017.

\bibitem{rethage2018eccv}
Dario Rethage, Johanna Wald, Jurgen Sturm, Nassir Navab, and Federico Tombari.
\newblock Fully-convolutional point networks for large-scale point clouds.
\newblock In {\em ECCV}, 2018.

\bibitem{le2018pointgrid}
Truc Le and Ye Duan.
\newblock Pointgrid: A deep network for 3d shape understanding.
\newblock In {\em CVPR}, 2018.

\bibitem{liu2019point}
Zhijian Liu, Haotian Tang, Yujun Lin, and Song Han.
\newblock Point-voxel cnn for efficient 3d deep learning.
\newblock In {\em NeurIPS}, 2019.

\bibitem{zhang2020deep}
Feihu Zhang, Jin Fang, Benjamin Wah, and Philip Torr.
\newblock Deep fusionnet for point cloud semantic segmentation.
\newblock In {\em ECCV}, 2020.

\bibitem{li2018pointcnn}
Yangyan Li, Rui Bu, Mingchao Sun, Wei Wu, Xinhan Di, and Baoquan Chen.
\newblock Pointcnn: Convolution on x-transformed points.
\newblock In {\em NeurIPS}, 2018.

\bibitem{li2018so}
Jiaxin Li, Ben~M Chen, and Gim~Hee Lee.
\newblock So-net: Self-organizing network for point cloud analysis.
\newblock In {\em CVPR}, 2018.

\bibitem{engelmann2018know}
Francis Engelmann, Theodora Kontogianni, Jonas Schult, and Bastian Leibe.
\newblock Know what your neighbors do: 3d semantic segmentation of point
  clouds.
\newblock In {\em ECCV}, 2018.

\bibitem{huang2018recurrent}
Qiangui Huang, Weiyue Wang, and Ulrich Neumann.
\newblock Recurrent slice networks for 3d segmentation of point clouds.
\newblock In {\em CVPR}, 2018.

\bibitem{jiang2019hierarchical}
Li Jiang, Hengshuang Zhao, Shu Liu, Xiaoyong Shen, Chi-Wing Fu, and Jiaya Jia.
\newblock Hierarchical point-edge interaction network for point cloud semantic
  segmentation.
\newblock In {\em ICCV}, 2019.

\bibitem{Yan_2020_CVPR}
Xu Yan, Chaoda Zheng, Zhen Li, Sheng Wang, and Shuguang Cui.
\newblock Pointasnl: Robust point clouds processing using nonlocal neural
  networks with adaptive sampling.
\newblock In {\em CVPR}, 2020.

\bibitem{fan2021scf}
Siqi Fan, Qiulei Dong, Fenghua Zhu, Yisheng Lv, Peijun Ye, and Fei-Yue Wang.
\newblock Scf-net: Learning spatial contextual features for large-scale point
  cloud segmentation.
\newblock In {\em CVPR}, 2021.

\bibitem{lian2020large}
Yanchao Lian, Tuo Feng, Jinliu Zhou, Meixia Jia, Aijin Li, Zhaoyang Wu, Licheng
  Jiao, Myron Brown, Gregory Hager, Naoto Yokoya, et~al.
\newblock Large-scale semantic 3-d reconstruction: Outcome of the 2019 ieee
  grss data fusion contest—part b.
\newblock {\em IEEE Journal of Selected Topics in Applied Earth Observations
  and Remote Sensing}, 14:1158--1170, 2020.

\bibitem{simonovsky2017dynamic}
Martin Simonovsky and Nikos Komodakis.
\newblock Dynamic edge-conditioned filters in convolutional neural networks on
  graphs.
\newblock In {\em CVPR}, 2017.

\bibitem{shen2018mining}
Yiru Shen, Chen Feng, Yaoqing Yang, and Dong Tian.
\newblock Mining point cloud local structures by kernel correlation and graph
  pooling.
\newblock In {\em CVPR}, 2018.

\bibitem{wang2018local}
Chu Wang, Babak Samari, and Kaleem Siddiqi.
\newblock Local spectral graph convolution for point set feature learning.
\newblock In {\em ECCV}, 2018.

\bibitem{landrieu2019point}
Loic Landrieu and Mohamed Boussaha.
\newblock Point cloud oversegmentation with graph-structured deep metric
  learning.
\newblock In {\em CVPR}, 2019.

\bibitem{wang2019graph}
Lei Wang, Yuchun Huang, Yaolin Hou, Shenman Zhang, and Jie Shan.
\newblock Graph attention convolution for point cloud semantic segmentation.
\newblock In {\em CVPR}, 2019.

\bibitem{chen2019clusternet}
Chao Chen, Guanbin Li, Ruijia Xu, Tianshui Chen, Meng Wang, and Liang Lin.
\newblock Clusternet: Deep hierarchical cluster network with rigorously
  rotation-invariant representation for point cloud analysis.
\newblock In {\em CVPR}, 2019.

\bibitem{li2019deepgcns}
Guohao Li, Matthias Muller, Ali Thabet, and Bernard Ghanem.
\newblock Deepgcns: Can gcns go as deep as cnns?
\newblock In {\em ICCV}, 2019.

\bibitem{wang2019dynamic}
Yue Wang, Yongbin Sun, Ziwei Liu, Sanjay~E Sarma, Michael~M Bronstein, and
  Justin~M Solomon.
\newblock Dynamic graph cnn for learning on point clouds.
\newblock {\em IEEE TOG}, 38(5):1--12, 2019.

\bibitem{su2018splatnet}
Hang Su, Varun Jampani, Deqing Sun, Subhransu Maji, Evangelos Kalogerakis,
  Ming-Hsuan Yang, and Jan Kautz.
\newblock Splatnet: Sparse lattice networks for point cloud processing.
\newblock In {\em CVPR}, 2018.

\bibitem{lei2019octree}
Huan Lei, Naveed Akhtar, and Ajmal Mian.
\newblock Octree guided cnn with spherical kernels for 3d point clouds.
\newblock In {\em CVPR}, 2019.

\bibitem{komarichev2019cnn}
Artem Komarichev, Zichun Zhong, and Jing Hua.
\newblock A-cnn: Annularly convolutional neural networks on point clouds.
\newblock In {\em CVPR}, 2019.

\bibitem{lan2019modeling}
Shiyi Lan, Ruichi Yu, Gang Yu, and Larry~S Davis.
\newblock Modeling local geometric structure of 3d point clouds using geo-cnn.
\newblock In {\em CVPR}, 2019.

\bibitem{mao2019interpolated}
Jiageng Mao, Xiaogang Wang, and Hongsheng Li.
\newblock Interpolated convolutional networks for 3d point cloud understanding.
\newblock In {\em ICCV}, 2019.

\bibitem{xie2018attentional}
Saining Xie, Sainan Liu, Zeyu Chen, and Zhuowen Tu.
\newblock Attentional shapecontextnet for point cloud recognition.
\newblock In {\em CVPR}, 2018.

\bibitem{hu2021towards}
Qingyong Hu, Bo Yang, Sheikh Khalid, Wen Xiao, Niki Trigoni, and Andrew
  Markham.
\newblock Towards semantic segmentation of urban-scale 3d point clouds: A
  dataset, benchmarks and challenges.
\newblock In {\em CVPR}, 2021.

\bibitem{fan2021pstnet}
Hehe Fan, Xin Yu, Yuhang Ding, Yi Yang, and Mohan Kankanhalli.
\newblock Pstnet: Point spatio-temporal convolution on point cloud sequences.
\newblock In {\em ICLR}, 2021.

\bibitem{duerr2020lidar}
Fabian Duerr, Mario Pfaller, Hendrik Weigel, and J{\"u}rgen Beyerer.
\newblock Lidar-based recurrent 3d semantic segmentation with temporal memory
  alignment.
\newblock In {\em 3DV}, 2020.

\bibitem{shi2020spsequencenet}
Hanyu Shi, Guosheng Lin, Hao Wang, Tzu-Yi Hung, and Zhenhua Wang.
\newblock Spsequencenet: Semantic segmentation network on 4d point clouds.
\newblock In {\em CVPR}, 2020.

\bibitem{zhang2020fusion}
Jiazhao Zhang, Chenyang Zhu, Lintao Zheng, and Kai Xu.
\newblock Fusion-aware point convolution for online semantic 3d scene
  segmentation.
\newblock In {\em CVPR}, 2020.

\bibitem{zhou2021tempnet}
Yunsong Zhou, Hongzi Zhu, Chunqin Li, Tiankai Cui, Shan Chang, and Minyi Guo.
\newblock Tempnet: Online semantic segmentation on large-scale point cloud
  series.
\newblock In {\em ICCV}, 2021.

\bibitem{chen2020improved}
Xinlei Chen, Haoqi Fan, Ross Girshick, and Kaiming He.
\newblock Improved baselines with momentum contrastive learning.
\newblock {\em arXiv preprint arXiv:2003.04297}, 2020.

\bibitem{chen2020simple}
Ting Chen, Simon Kornblith, Mohammad Norouzi, and Geoffrey Hinton.
\newblock A simple framework for contrastive learning of visual
  representations.
\newblock In {\em ICML}, 2020.

\bibitem{he2020momentum}
Kaiming He, Haoqi Fan, Yuxin Wu, Saining Xie, and Ross Girshick.
\newblock Momentum contrast for unsupervised visual representation learning.
\newblock In {\em CVPR}, 2020.

\bibitem{yin2022proposalcontrast}
Junbo Yin, Dingfu Zhou, Liangjun Zhang, Jin Fang, Cheng-Zhong Xu, Jianbing
  Shen, and Wenguan Wang.
\newblock Proposalcontrast: Unsupervised pre-training for lidar-based 3d object
  detection.
\newblock In {\em ECCV}, 2022.

\bibitem{dosovitskiy2015discriminative}
Alexey Dosovitskiy, Philipp Fischer, Jost~Tobias Springenberg, Martin
  Riedmiller, and Thomas Brox.
\newblock Discriminative unsupervised feature learning with exemplar
  convolutional neural networks.
\newblock {\em IEEE TPAMI}, 38(9):1734--1747, 2015.

\bibitem{caron2020unsupervised}
Mathilde Caron, Ishan Misra, Julien Mairal, Priya Goyal, Piotr Bojanowski, and
  Armand Joulin.
\newblock Unsupervised learning of visual features by contrasting cluster
  assignments.
\newblock In {\em NeurIPS}, 2020.

\bibitem{gutmann2010noise}
Michael Gutmann and Aapo Hyv{\"a}rinen.
\newblock Noise-contrastive estimation: A new estimation principle for
  unnormalized statistical models.
\newblock In {\em AISTATS}, 2010.

\bibitem{oord2018representation}
Aaron van~den Oord, Yazhe Li, and Oriol Vinyals.
\newblock Representation learning with contrastive predictive coding.
\newblock {\em arXiv preprint arXiv:1807.03748}, 2018.

\bibitem{hjelm2019learning}
R~Devon Hjelm, Alex Fedorov, Samuel Lavoie-Marchildon, Karan Grewal, Phil
  Bachman, Adam Trischler, and Yoshua Bengio.
\newblock Learning deep representations by mutual information estimation and
  maximization.
\newblock In {\em ICLR}, 2019.

\bibitem{xie2020pointcontrast}
Saining Xie, Jiatao Gu, Demi Guo, Charles~R Qi, Leonidas Guibas, and Or Litany.
\newblock Pointcontrast: Unsupervised pre-training for 3d point cloud
  understanding.
\newblock In {\em ECCV}, 2020.

\bibitem{xie2021propagate}
Zhenda Xie, Yutong Lin, Zheng Zhang, Yue Cao, Stephen Lin, and Han Hu.
\newblock Propagate yourself: Exploring pixel-level consistency for
  unsupervised visual representation learning.
\newblock In {\em CVPR}, 2021.

\bibitem{wang2021dense}
Xinlong Wang, Rufeng Zhang, Chunhua Shen, Tao Kong, and Lei Li.
\newblock Dense contrastive learning for self-supervised visual pre-training.
\newblock In {\em CVPR}, 2021.

\bibitem{wang2021exploring}
Wenguan Wang, Tianfei Zhou, Fisher Yu, Jifeng Dai, Ender Konukoglu, and Luc
  Van~Gool.
\newblock Exploring cross-image pixel contrast for semantic segmentation.
\newblock In {\em ICCV}, 2021.

\bibitem{zhou2022rethinking}
Tianfei Zhou, Wenguan Wang, Ender Konukoglu, and Luc Van~Gool.
\newblock Rethinking semantic segmentation: A prototype view.
\newblock In {\em CVPR}, 2022.

\bibitem{lianggmmseg}
Chen Liang, Wenguan Wang, Jiaxu Miao, and Yi Yang.
\newblock Gmmseg: Gaussian mixture based generative semantic segmentation
  models.
\newblock In {\em NeurIPS}, 2022.

\bibitem{jiang2021guided}
Li Jiang, Shaoshuai Shi, Zhuotao Tian, Xin Lai, Shu Liu, Chi-Wing Fu, and Jiaya
  Jia.
\newblock Guided point contrastive learning for semi-supervised point cloud
  semantic segmentation.
\newblock In {\em ICCV}, 2021.

\bibitem{yin2022semi}
Junbo Yin, Jin Fang, Dingfu Zhou, Liangjun Zhang, Cheng-Zhong Xu, Jianbing
  Shen, and Wenguan Wang.
\newblock Semi-supervised 3d object detection with proficient teachers.
\newblock In {\em ECCV}, 2022.

\bibitem{meng2021towards}
Qinghao Meng, Wenguan Wang, Tianfei Zhou, Jianbing Shen, Yunde Jia, and Luc
  Van~Gool.
\newblock Towards a weakly supervised framework for 3d point cloud object
  detection and annotation.
\newblock {\em IEEE TPAMI}, 2021.

\bibitem{xie2016unsupervised}
Junyuan Xie, Ross Girshick, and Ali Farhadi.
\newblock Unsupervised deep embedding for clustering analysis.
\newblock In {\em ICML}, 2016.

\bibitem{yang2016joint}
Jianwei Yang, Devi Parikh, and Dhruv Batra.
\newblock Joint unsupervised learning of deep representations and image
  clusters.
\newblock In {\em CVPR}, 2016.

\bibitem{caron2018deep}
Mathilde Caron, Piotr Bojanowski, Armand Joulin, and Matthijs Douze.
\newblock Deep clustering for unsupervised learning of visual features.
\newblock In {\em ECCV}, 2018.

\bibitem{ji2019invariant}
Xu Ji, Joao~F Henriques, and Andrea Vedaldi.
\newblock Invariant information clustering for unsupervised image
  classification and segmentation.
\newblock In {\em ICCV}, 2019.

\bibitem{zhou2021group}
Tianfei Zhou, Liulei Li, Xueyi Li, Chun-Mei Feng, Jianwu Li, and Ling Shao.
\newblock Group-wise learning for weakly supervised semantic segmentation.
\newblock {\em IEEE Transactions on Image Processing}, 31:799--811, 2021.

\bibitem{li2020prototypical}
Junnan Li, Pan Zhou, Caiming Xiong, and Steven Hoi.
\newblock Prototypical contrastive learning of unsupervised representations.
\newblock In {\em ICLR}, 2020.

\bibitem{li2021contrastive}
Yunfan Li, Peng Hu, Zitao Liu, Dezhong Peng, Joey~Tianyi Zhou, and Xi Peng.
\newblock Contrastive clustering.
\newblock In {\em AAAI}, 2021.

\bibitem{liang2023clustseg}
James Liang, Tianfei Zhou, Dongfang Liu, and Wenguan Wang.
\newblock Clustseg: Clustering for universal segmentation.
\newblock {\em ICML}, 2023.

\bibitem{wang2022looking}
Wenguan Wang, Guolei Sun, and Luc Van~Gool.
\newblock Looking beyond single images for weakly supervised semantic
  segmentation learning.
\newblock {\em IEEE TPAMI}, 2022.

\bibitem{berman2018lovasz}
Maxim Berman, Amal~Rannen Triki, and Matthew~B Blaschko.
\newblock The lov{\'a}sz-softmax loss: A tractable surrogate for the
  optimization of the intersection-over-union measure in neural networks.
\newblock In {\em CVPR}, 2018.

\bibitem{wang2020cross}
Xun Wang, Haozhi Zhang, Weilin Huang, and Matthew~R Scott.
\newblock Cross-batch memory for embedding learning.
\newblock In {\em CVPR}, 2020.

\bibitem{tishby2015deep}
Naftali Tishby and Noga Zaslavsky.
\newblock Deep learning and the information bottleneck principle.
\newblock In {\em IEEE Information Theory Workshop}, pages 1--5, 2015.

\bibitem{yan2021sparse}
Xu Yan, Jiantao Gao, Jie Li, Ruimao Zhang, Zhen Li, Rui Huang, and Shuguang
  Cui.
\newblock Sparse single sweep lidar point cloud segmentation via learning
  contextual shape priors from scene completion.
\newblock In {\em AAAI}, 2021.

\bibitem{cheng20212}
Ran Cheng, Ryan Razani, Ehsan Taghavi, Enxu Li, and Bingbing Liu.
\newblock (af)2-s3net: Attentive feature fusion with adaptive feature selection
  for sparse semantic segmentation network.
\newblock In {\em CVPR}, 2021.

\bibitem{xu2021rpvnet}
Jianyun Xu, Ruixiang Zhang, Jian Dou, Yushi Zhu, Jie Sun, and Shiliang Pu.
\newblock Rpvnet: A deep and efficient range-point-voxel fusion network for
  lidar point cloud segmentation.
\newblock In {\em ICCV}, 2021.

\bibitem{hou2022point}
Yuenan Hou, Xinge Zhu, Yuexin Ma, Chen~Change Loy, and Yikang Li.
\newblock Point-to-voxel knowledge distillation for lidar semantic
  segmentation.
\newblock In {\em CVPR}, 2022.

\bibitem{qiu2021semantic}
Shi Qiu, Saeed Anwar, and Nick Barnes.
\newblock Semantic segmentation for real point cloud scenes via bilateral
  augmentation and adaptive fusion.
\newblock In {\em CVPR}, 2021.

\bibitem{lu2021cga}
Tao Lu, Limin Wang, and Gangshan Wu.
\newblock Cga-net: Category guided aggregation for point cloud semantic
  segmentation.
\newblock In {\em CVPR}, 2021.

\bibitem{tang2022contrastive}
Liyao Tang, Yibing Zhan, Zhe Chen, Baosheng Yu, and Dacheng Tao.
\newblock Contrastive boundary learning for point cloud segmentation.
\newblock In {\em CVPR}, 2022.

\bibitem{lai2022stratified}
Xin Lai, Jianhui Liu, Li Jiang, Liwei Wang, Hengshuang Zhao, Shu Liu, Xiaojuan
  Qi, and Jiaya Jia.
\newblock Stratified transformer for 3d point cloud segmentation.
\newblock In {\em CVPR}, 2022.

\bibitem{wupoint}
Xiaoyang Wu, Yixing Lao, Li Jiang, Xihui Liu, and Hengshuang Zhao.
\newblock Point transformer v2: Grouped vector attention and partition-based
  pooling.
\newblock In {\em NeurIPS}, 2022.

\bibitem{Lang_2019_CVPR}
Alex~H. Lang, Sourabh Vora, Holger Caesar, Lubing Zhou, Jiong Yang, and Oscar
  Beijbom.
\newblock Pointpillars: Fast encoders for object detection from point clouds.
\newblock In {\em CVPR}, 2019.

\bibitem{alonso2020MiniNet3D}
I{\~n}igo Alonso, Luis Riazuelo, Luis Montesano, and Ana~C Murillo.
\newblock 3d-mininet: Learning a 2d representation from point clouds for fast
  and efficient 3d lidar semantic segmentation.
\newblock In {\em IROS}, 2020.

\end{thebibliography}
}

\end{document}